\documentclass{article}

\usepackage{arxiv}

\usepackage{amssymb}
\usepackage{amsmath}
\usepackage{hyperref}
\usepackage{xcolor}
\usepackage{graphicx}
\usepackage{bbm}
\usepackage{multirow}
\usepackage{cite}
\usepackage{booktabs} 

\usepackage[noend]{algpseudocode}
\usepackage[ruled,vlined,linesnumbered,noend]{algorithm2e}

\definecolor{green}{rgb}{0.0, 0.6, 0.0}
\usepackage{pifont}
\newcommand{\cmark}{\ding{51}}%
\newcommand{\xmark}{\ding{55}}%

\newcommand{\modif}[1]{{#1}}

\newcommand{\seqUser}{{s_u}}
\newcommand{\groundTruth}{{s_u^t}}
\newcommand{\groundTruthBis}{{g_u}}
\newcommand{\RankingGroundTruthBis}{{R_{g_u}}}
\newcommand{\seqUserToT}{{s_u^{\lbrack 1,t\rbrack}}}
\newcommand{\seqUserToTMinusOne}{{s_u^{\lbrack 1,t\lbrack}}}
\newcommand{\seqUserXToTMinusOne}{{s_u^{\lbrack x,t\lbrack}}}

\newcommand{\setUser}{{I_{\seqUser}}}
\newcommand{\setUserToT}{{I_{\seqUserToT}}}
\newcommand{\setUserToTMinusOne}{{I_{\seqUserToTMinusOne}}}
\newcommand{\setUserXToTMinusOne}{{I_{\seqUserXToTMinusOne}}}

\newcommand{\freq}{{{F}}}
\newcommand{\freqUser}{{m_{\seqUser}}}
\newcommand{\freqUserToT}{{m_{\seqUserToT}}}
\newcommand{\freqUserToTMinusOne}{{m_{\seqUserToTMinusOne}}}
\newcommand{\RItemFreqUserToTMinusOne}{{m^{r}_{\seqUserToTMinusOne}}}
\newcommand{\frequ}{{ m_{s_u^{\lbrack 1,t\lbrack}}}}

\newcommand{\predItemIJ}{{\widehat{p}_{i|j}}}
\newcommand{\predUserItem}{{\widehat{p}_{u,i}}}
\newcommand{\predUserItemIJ}{{\widehat{p}_{u,i|j}}}
\newcommand{\predUserItemTime}{{\widehat{p}_{u,i,t}}}
\newcommand{\predUserItemJTime}{{\widehat{p}_{u,j,t}}}
\newcommand{\predUsertemTruthTime}{{\widehat{p}_{u,\groundTruth,t}}}
	
\newcommand{\TotalOrder}{{\: >_{u,t}\:}}
\newcommand{\MaxLenghtUP}{{\textsc{max\_length}}}
\newcommand{\minCount}{{\textsc{minCount}}}
\newcommand{\maxSize}{{\textsc{L}}}

\newcommand{\algo}{{{\bf REBUS}}}
\newcommand{\algoST}{{{\bf REBUS\_SD}}}
\newcommand{\algoLT}{{{\bf REBUS\_UP}}}
\newcommand{\algolong}{{{Recommendation Embedding Based on freqUent Sequences}}}

\newcommand{\urlCode}{{\url{https://bit.ly/2PKVpC2}}}

\newcommand{\ptitle}[1]{\vspace{1mm}\noindent{\bf #1.}}

\begin{document}

\title{Sequential recommendation with metric models based on frequent sequences}
		
\author{
	\textbf{Corentin Lonjarret}\\ INSA Lyon, CNRS, LIRIS UMR5205\\ \texttt{corentin.lonjarret@insa-lyon.fr} \and
	\textbf{Roch Auburtin} \\  Visiativ, France\\ \texttt{rauburtin@visiativ.com}\\ \and
	\textbf{C\'eline Robardet} \\ INSA Lyon, CNRS, LIRIS UMR5205\\ \texttt{celine.robardet@insa-lyon.fr} \and 
	\textbf{Marc Plantevit} \\ Universit\'e Lyon1, CNRS, LIRIS UMR5205\\ \texttt{marc.plantevit@liris.cnrs.fr} 
}


\maketitle
            
\begin{abstract}
  Modeling user preferences (long-term history) and user dynamics (short-term history) is of greatest importance to build efficient
  sequential recommender systems. The challenge lies in the successful combination of the whole user's history and his recent actions
  (sequential dynamics) to provide personalized recommendations. Existing methods capture the sequential dynamics of
  a user using fixed-order Markov chains (usually first order chains) regardless of the user, which limits both the impact of the past of
  the user on the recommendation and the ability to adapt its length to the user profile. In this article, we propose to use frequent
  sequences to identify the most relevant part of the user history for the recommendation. The most salient items are then used in a
  unified metric model that embeds items based on user preferences and sequential dynamics.
  Extensive experiments demonstrate that our method outperforms state-of-the-art, especially on sparse datasets. 
  We show that considering sequences of varying lengths improves the recommendations and we also emphasize that these
  sequences provide explanations on the recommendation.\\ 
  
\end{abstract}

\keywords{Recommender systems \and Sequential Recommendation \and Collaborative Filtering}






\section{Introduction}

As society becomes ever more digital, the number of situations where users are faced with a huge number of choices increases. In this context, a selection process by exhaustive examination of all possibilities is not feasible. Hence, accessing a virtually infinite set of digital contents -- or digital descriptions of real objects called hereafter items -- is only possible if the search process is equipped with powerful recommendation tools. This is what recommender systems \cite{BookSR,gomez2016netflix,resnick1997recommender} do: Exploiting the traces left by users to recommend a small set of items of interest to them.

In this paper, we tackle the problem of sequential recommendation that aims to predict/recommend to a user the next item from collaborative data. This is a challenging problem whose importance has been gradually recognized by researchers. Indeed, user preferences (i.e., the long-term dynamics) and user sequential dynamics (i.e., the short-term dynamics) need to be fruitfully combined to account for both personalization and sequential transitions.
The first methods, which paved this research field, aim at capturing either user preferences or user sequential dynamics. User preferences have been identified using methods such as Matrix Factorization (MF) \cite{CFT,MF} that uses inner products to decompose a compatibility matrix between users and items, or FISM \cite{FISM} that splits a similarity matrix between items to better account for transitive relationships between items. Sequential dynamics has been caught by Markov Chains (MC) for the calculation of the conditional probability of appearance of an item according to a fixed number $L $ of passed items.
The next generation of recommender systems is based on models that combine both sequential dynamics and user preferences, like FPMC \cite{FPMC} or FOSSIL \cite{FOSSIL} which fuse Matrix Factorization and Markov Chains. PRME \cite{PRME} improves FOSSIL and FPMC by replacing inner product with Euclidean distance. Such a metric embedding  brings a better generalization, mostly due to the respect of the triangle inequality. However, it still suffers from a lack of personalization \modif{on the short term part}, for example due to the fact that Markov chains have a fixed length regardless of the users and their considered items.
Moreover, these models combine the user preferences and sequential dynamics using two embeddings -- one for the user preferences and another one for the sequential dynamics -- which can damage the quality of the model.
More recently, and especially since the traces left by the users become longer, a growing research effort \cite{DeepLearningSurvey,CASER,RUM:WSDM18} has been dedicated to the investigation of the use of deep learning technique for sequential recommendation.
Yet, this kind of methods need large amounts of data and can have troubles on sparse datasets. These statements are confirmed by \modif{our} experiments.

To cope with these limitations, we propose a new model \algo{} (\algolong) that uses (1) frequent sequences to identify the part of user history that is the most relevant for recommendation and using these sequences to estimate Markov Chains of variable orders and (2) a unified metric embedding model based on user preferences and user sequential dynamics.

\begin{figure}[htb]
	\includegraphics[width=1\textwidth]{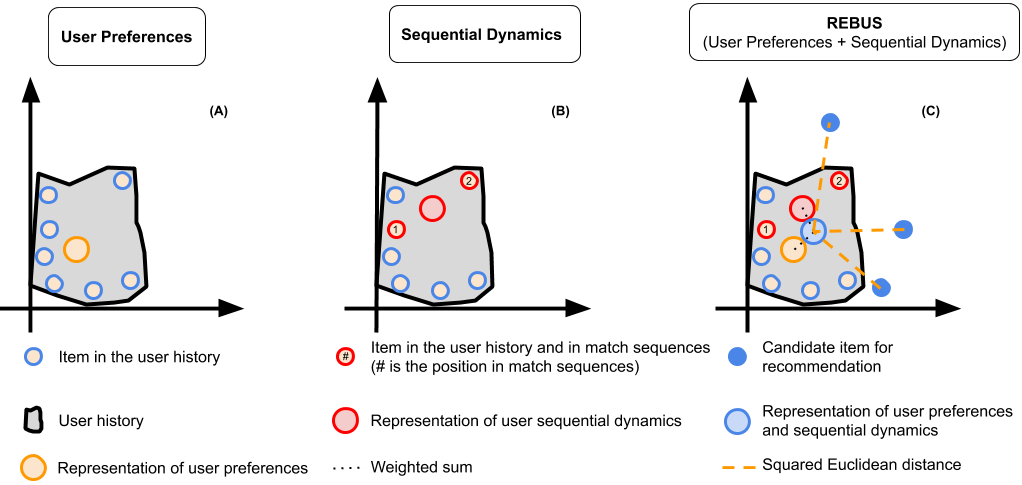}
	\caption{Overview of \algo{} that accommodates user preferences and sequential dynamics through Euclidean distances. (a) \textbf{User preferences} \modif{are} represented by the embedding of all the user's history items. (b) \textbf{Sequential Dynamics} is represented by the embedding of the user's items that match a frequent sequence. The temporal order of the items is taken into account with a temporal damping factor. (c) \textbf{\algo} takes the nearest item of the weighted sum of user preferences and sequential dynamics for recommendation.}
	\label{fig:overview}
\end{figure}

Figure \ref{fig:overview} illustrates how \algo{} works. It consists of embedding items into latent vectors and using these vectors to represent user preferences and user  sequential dynamics: In Fig.~\ref{fig:overview}(a), user preferences \modif{are} represented by the weighted sum of all the items of the user history and in Fig.~\ref{fig:overview}(b) user sequential dynamics is represented by the weighted sum of the most representative recent items of the user history. These most representative items are those that match frequent sequences over the entire dataset. These frequent sequences are used as a proxy to know which items in the user history are important to capture sequential dynamics. In Figure~\ref{fig:overview}(c), \algo{} recommend the item that is the nearest to the vector resulting from the weighted sum of user preferences and sequential dynamics.

Our contributions are summarized as follows. We develop a new method, \algo, that integrates and unifies metric embedding models to capture both user preferences and sequential dynamics. Especially, the use of personalized sequential patterns makes it possible to select the part of the recent user's history that is of most interest for the recommendation. In an empirical study over \modif{12} datasets, we show that \algo{} outperforms \modif{9} state-of-the-art algorithms (that cover the different types introduced above) on sparse dataset and have nearly the same performance on dense datasets. Furthermore, we show that the sequences we use to model user sequential dynamics can provide additional insights on the recommendations. Finally, for reproducibility purposes, data and code are made available here: \urlCode{}.

The remaining of the paper is organized as follows. Section \ref{sec:notation} introduces the notations. We review the literature in Section \ref{sec:rw}. \algo{} model is defined in Section \ref{sec:model}. We report an extensive empirical study in Section \ref{sec:xp} and conclude in Section \ref{sec:con}.

\section{Notations}\label{sec:notation}
In this section, we introduce the notations we use throughout this paper to both define our proposal and review the literature.  

$U$ and $I$ are used to respectively denote the set of users and the set of items. The symbols $u$ and $i$ stand for individual users and items (i.e., $u\in U$ and $i\in I$). The trace left by a user $u$ is the sequence of items $\seqUser{}=\langle i_{1},\dots, i_p\rangle$, with $i_\ell\in I,\: \ell=1\dots p$, and $i_p$ being the most recent item. Let $\seqUserToT{}$ be the substring of $\seqUser{}$ starting at the $1^{st}$ and oldest item, and ending at the $t^{th}$ and more recent one. We use $\setUser{}$ to denote items that appear in $\seqUser{}$ and $\setUserToT{}$ for those that belong to $\seqUserToT{}$. Notations used throughout this paper are summarised and exemplified in Table~\ref{tab:notation}.

\begin{table}[htb]
  \begin{center}
    \caption{Notations.}
    \label{tab:notation}
    \scalebox{1}{
      \begin{tabular}{|p{1cm}|p{10cm}|}
	\hline
	$U$ & User set.\\ \hline
	$I$ & Item set. \\ \hline
	$S$ & Set of user sequences.\\ \hline
	$\freq{}$ & Set of frequent substrings of $S$.  $\freq{} = \{ \langle1\rangle, \langle2\rangle, \langle4\rangle,$ $ \langle5\rangle, \langle4,5\rangle\} $\\\hline
	$\seqUser{}$ & Sequence of items associated to user $u$.  $\seqUser{}= \langle 1,2,3,2,2,$ $4,5\rangle$\\\hline
	$\setUser{}$ & Set of items that appear in $\seqUser{}$. $\setUser{} = \{1,2,3,4,5\}$, \quad $I_{\seqUser{}^{[1,4]}}  = \{1,2,3\}$\\\hline
	$\freqUser{}$ & Sequence of $\freq{}$ that best matches $\seqUser{}$. $\freqUser{} = \langle4,5\rangle $\\\hline
	$\seqUser{}^t$ & Item that appears in $\seqUser{}$ at position $t$. $\seqUser{}^{5} = 2$\\ \hline
	$\freqUser{}^t$ & Item that appears in $\freqUser{}$ at position $t$. $\freqUser{}^2 = 5 $\\ \hline
	$\seqUserToTMinusOne{}$ & Substring of $\seqUser{}$ starting at position 1 and ending at position $t-1$.  $\seqUser{}^{[1,5[} = \langle 1,2,3,2\rangle$	\\\hline
	    $\freqUserToTMinusOne{}$ & Sequence of $\freq{}$ that best matches $\seqUserToTMinusOne{}$. $\freqUser{}^{[1,5[} = \langle 2\rangle$ 	\\\hline
		$\predUserItemTime{}$ & Prediction that measures the probability for user $u$ to choose item $i$ while only considering $\seqUserToTMinusOne{}$. \\ \hline
		$\TotalOrder{}$ & Personalized total order of user $u$ at \modif{position} $t$. \\
		\hline
      \end{tabular}
	      }
  \end{center}
\end{table}

\section{Related Work}\label{sec:rw}

In this paper, we consider {\em sequential recommendation} that aims to predict the next item that will interest a user based on a sequence of items that he previously purchased/consumed. 

There are several variants to this problem which lie outside the scope of this study. For instance, {\em temporal recommendation} \cite{TimeWeightCF, CFT, BPTF, RRN_Temporal} takes into account the timestamps of users' actions to make recommendations based on a specific time (e.g. recommending a glass of wine in the evening but not in the morning).
\modif{Also, we exclude {\em context-aware} \cite{libFM, TransFM, CKE, KSR} and {\em time-aware} \cite{BFUB} recommendations, as these approaches require contextual information (timestamps, item categories, user features, etc...) which limits their use to specific data. Hence, we restrain our study to methods that only take as input users' sequences to make recommendation, without considering additional information.}

\modif{In the following, we detail the related works that consider recommender systems as an estimation problem of $\widehat{p}_{u,i}$, values that measure} the interest of user $u$ for item $i$ using real observations $\bf{D}$. \modif{Such methods rely on the optimization of} a loss function $\mathcal{L({\bf D},{\bf \widehat{P}})}$, which evaluates the proximity between ${\bf D}$ and ${\bf \widehat{P}}=(\widehat{p}_{u,i})$. We present below the recent approaches that are related to our proposal. 

\subsection{User preferences (Long Term Dynamics)}
Early recommendation methods aim at identifying user preferences using Collaborative Filtering techniques such as K-Nearest Neighbour algorithms \cite{KNN} or Matrix Factorization (MF) \cite{MF}. In many state-of-the-art recommender systems, MF is used to model the interactions between users and items. It consists in decomposing a matrix, ${\bf D}_{\mbox{user}\times \mbox{item}}=(d_{u,i})$ with $d_{u,i}=1\Leftrightarrow i\in I_u$, in a product of $k$-ranked matrices: ${\bf D}_{\mbox{user}\times \mbox{item}}\approx R\times Q$. The prediction $\predUserItem{}$, that the user $u$ chooses the item $i$, is estimated by the inner product: 
$$\predUserItem{} \propto \langle \textbf{R}_{u}, \textbf{Q}_{i} \rangle ,$$ 
with ${\bf R}_{u}$ and ${\bf Q}_{i}$ the latent vectors associated to user $u$ and item $i$. However, as usually users give implicit feedback on few items of $I$, the matrix ${\bf D}$ is sparse and these approaches suffer from loss of precision when the numbers of users and items grow. In order to overcome these problems, other approaches such as FISM~\cite{FISM} -- which is an improvement of SLIM \cite{SLIM} -- decompose an implicit item-to-item similarity matrix in two $k$-ranked matrices ${\bf P}$ and ${\bf Q}$ so that:
$$ \predUserItem{} \propto \beta_{i}+\beta_u + \frac{1}{|\setUser{}	\backslash \{i\}|^{\alpha}} \sum_{j \in \setUser{} \setminus \{i\}} \langle {\bf P}_{j},{\bf Q}_{i}\rangle ,$$ where $\beta_i, \beta_u$ are biases respectively associated to item $i$ and user $u$, $\frac{1}{|\setUser{} \setminus \{i\}|}$ normalized long set of items and ${\alpha}$ is used to control the degree of agreement between items. Hence, the more $i$ is similar to items already chosen by user $u$, the more likely $i$ will be a good choice for $u$. Taking into account these transitive relations between items makes it possible to increase the quality of the recommendation.

\modif{Recently, deep learning techniques have been introduced in many recommender systems. For example, user preferences are modeled by Multi-Layer Perceptions (MLP) in \cite{NeuMF} or using Auto-encoder in \cite{AutoRec}.}

\subsection{Sequential Dynamics (Short Term Dynamics)}
Another trend in recommender system design is to use sequential information (the order in which the items are chosen) to model the user sequential dynamics. Short term dynamics can be modeled using first order Markov Chains. It consists in using the probability $p(i\mid j)$ of predicting $i$ given the last item $j$ in the sequence of the user. The transition matrix $\left( p(i\mid j)\right)_{ij}$ is decomposed in a product of two $k$-ranked matrices ${\bf M}_{j}$ and ${\bf N}_{i}$ that represent the latent/embedding vectors of item $j$ and $i$, so that the probability of having $i$ after $j$ is estimated by the inner product:
$$\predItemIJ{} \propto \langle {\bf M}_{j},{\bf N}_{i}\rangle,$$ 
with ${\bf M}_{j}$ and ${\bf M}_{i}$ the latent vectors associated to items $j$ and $i$. It is also possible to consider high order Markov Chains with the same principle.

\modif{
There are other approaches, called session-based model \cite{SequenceAwareRS, SBSR}, that give more weight to recent events by construction: The user sequences is divided in sessions. Therefore,  they focus on shorter user sequences (i.e., the sessions) fostering short term dynamics. Considering sessions makes it possible to focus the recommendation on similarities with recent items. Some of these methods, like \textit{Session-based recommendations with Recurrent Neural Networks} (GRU4Rec) \cite{GRU4REC} or Session-based KNN
\cite{sessionBased}, are widely used and deserve to be mentioned and evaluated within the framework of sequential recommendation systems.}

\subsection{Unifying user preferences and sequential dynamics}
Current recommendation methods accommodate user preferences and sequential dynamics as it has been observed that it increases their performances. FPMC \cite{FPMC} is one of the first method that uses both Matrix Factorization and first-order factorized Markov Chains. The probability that user $u$ chooses item $i$ just after having taken item $j$ is estimated by the sum of two inner products:
$$ \predUserItemIJ{} \propto \langle {\bf R}_{u},{\bf Q}_{i}\rangle +\langle {\bf M}_{j},{\bf N}_{i}\rangle$$

Fossil \cite{FOSSIL} improves FPMC by associating an implicit item-to-item similarity matrix decomposition method with a Markov Chains model of order $L$. On top of that, PRME \cite{PRME} enhances FPMC and Fossil by replacing the inner products with Euclidean distances. As argued in \cite{LME,PRME}, metric embedding model brings better generalization ability than Matrix Factorization to represent Markov chains because of the triangle inequality assumption. The probability that user \textit{u} takes the item \textit{i} after item \textit{j} is estimated by the sum of the following Euclidean distances:
$$ \predUserItemIJ{} \propto -\left( \alpha \cdot ||{\bf R}_{u} - {\bf Q}_{i}||_{2}^{2} + (1-\alpha) \cdot ||{\bf M}_{j} - {\bf M}_{i}||_{2}^{2}\right)$$
with $\alpha$ a weight that controls the long term and short term dynamics.

TransRec \cite{TransRec} is based on a novel metric embedding model that unifies user preferences and sequential dynamics with translation. To achieve this, items are embedded as points $P_i$ in a latent transition space and users are modeled as translation vectors $T_u$ in the same space:
$$\predUserItemIJ{} \propto \beta_{i} - d(P_{j} + T_{u}, P_{i})$$
with $d()$ a distance ($L_{1}$ or squared $L_{2}$), $T_{u}$ a translation vector representing $u$, and $P_{i}$, $P_{j}$ points in the transition space related to items $i$ and $j$. The sequential dynamics is captured with first order Markov chains.

Recently, a growing research effort has been dedicated to the investigation of the use of deep learning models \cite{DeepLearningSurvey} for sequential recommendation. In that context, Recurrent Neural Networks (RNN) have been widely used for session-based recommendation. Such networks use Long Short Term Memory (LSTM) or Gated Recurrent Units (GRU) and have shown good results to model user sequential Dynamics \cite{GRU4REC,GRU4REC2,LSR}. Another line of work has investigated Convolutional Neural Network based methods. Convolutional Sequence Embedding Recommendation Model (Caser) \cite{CASER} is the first CNN-based method that captures both user preferences and user sequential dynamics. Caser embeds $L$ previously considered items as an "image" and learns sequential patterns with convolution operations. Inspired by a new sequential model {\em Transformer} for machine translation tasks \cite{SelfAttention}, Self-Attentive Sequential Recommendation (SASRec) \cite{SASRec} is a new sequential recommendation model that outperforms many advanced sequential models on sparse and dense datasets.

\modif{Finally, we can cite the recent work of \cite{BFUB}, which, alike to our approach, computes a similarity based on the longest common subsequence. However, this method requires temporal information, which makes it out of the scope of our study.}

\subsection{Summary and desiderata} Few approaches combine user preferences and sequential dynamics into a metric embedding model. The proposed approaches consider Markov chains of fixed small order $L$ or black box model based on deep neural networks. In this paper, we aim to personalize the sequential dynamics of users by identifying the most salient items from the user's history, and use them with user preferences in order to have a unified embedding metric model. Our goal is also to have a model that is as interpretable as possible and gives some explanations on the recommendation made.

\section{\algo{} model} \label{sec:model}
\algo{} is a metric embedding model in which only items are projected. Their corresponding embedding vectors are influenced by both the preferences of the user and their sequential dynamics. The user preferences \modif{are} wrapped in the model by constraining the latent vector  $P_i$ of an item that should be recommended to a user to be as close as possible to the average vector of the embedding of the items contained in the user history. 
In order to identify the part of the sequence that is most characteristic of a user, we consider frequent sequential patterns. These frequent sequences are both present in the history of several users (whose minimum number is specified by a threshold) while allowing to ignore certain items which are not sufficiently characteristic of their general behavior. Once the most important items are identified for a user, their order is taken into account in the model thanks to a damping factor based on the rank of the item in the sequence. This allows sequences of different lengths to be used in a unified manner. The embedding vectors are then learned using the Bayesian personalized ranking optimization method \cite{BPR} on our model. These steps are detailed below.

\subsection{Long-term metric-based model}
To model user preferences, we follow the way paved by FISM \cite{FISM} and FOSSIL \cite{FOSSIL} while replacing inner product with Euclidean distance. The objective is to compute a latent vector $P_i$ for each item $i$, so that the prediction for a user $u$ to choose $i$ varies in the opposite way to the distance between the sum of the items already chosen by $u$ and the item $i$. Moreover, in order to not overweight items that appear in long transactions -- items selected with many others -- we normalize the user preferences by the inverse of the number of items in the sequence of user $u$. As in FISM or FOSSIL, the hyperparameter $\alpha$ controls the degree of agreement between items:
When $\alpha = 1$, the long-term part is equivalent to the barycenter of the latent vectors; The closer $\alpha$ is to 0, the more it is equivalent to the sum of latent vectors. In \modif{our} experiments, we observed that the best values of $\alpha$ lie between 0.7 and 1.
\begin{eqnarray*}
	\predUserItem ~\propto~ -|| \frac{1}{|\setUser \setminus \{i\}|^{\alpha}} \sum_{j \in \setUser \backslash \{i\}} P_{j} - P_{i} ||_{2}^{2}.
\end{eqnarray*}

In the learning phase, we estimate the prediction associated to item $i$ and user $u$ at \modif{position $t$, where $i$ is the $t^{th}$ item taken by $u$. Note that, for a user $u$, the first item is the most older one ($t=1$) and the last taken item is the most recent one ($t=|\seqUser{}|$)}.
It seems also rightful to restrict the set of items to be considered to those older than $t$ ($\setUserToTMinusOne$) (even if it is not the choice made by FISM and FOSSIL).
This choice was confirmed empirically. In addition, we limit the number of items to be considered to the recent ones with $\MaxLenghtUP$ hyperparameter. It allows \algo{} to be more flexible by controlling the temporal window that influences the user's preferences and to get rid of the old past. All in one, this leads to the following equation for estimating user long-term preferences:
\begin{eqnarray}
\label{eq1}
\predUserItemTime \propto -|| \frac{1}{|\setUserXToTMinusOne \setminus \{i\}|^{\alpha}} \sum_{j \in \setUserXToTMinusOne \backslash \{i\}} P_{j} - P_{i} ||_{2}^{2},
\end{eqnarray}
with $x=\max\left(t - \MaxLenghtUP,1\right)$.

\subsection{Short-term dynamics modeled  by frequent patterns}
As explained in Section~\ref{sec:rw}, it has been shown that taking into account short-term individual dynamics improves the recommendation. However, existing approaches only consider a fixed, short and consecutive part of the history to make the recommendation. In contrast, \algo{} takes into account parts of the user history which may be of different lengths for each user and also not necessarily consecutive. Finding the most adapted sequential pattern for a user $u$ at a \modif{position} $t$ is accomplished in two steps: (1) computing a set $F$ of representative sequences of  users' histories, and (2) identifying a sequence that personally represents a user $u$.

However, taking into account inter-individual variability -- by allowing items to be   in the representative sequence of a user -- can be done either in step 1 or in step 2. In the first scenario, we extract \textit{frequent subsequences} from user's histories, and identify the subsequence that \textit{perfectly matches} the history of a given
user when making the recommendation. In the second scenario, we extract \textit{frequent substrings} from user's histories and identify the substring that best characterizes a user using a string alignment algorithm that allows to skip some uninformative items.

We implemented both scenarios and found that the quality of the results are very similar. However, the model based on frequent subsequences has a higher cost due to the size of the collection of frequent subsequences, that is much greater than the one of frequent substrings. We thus detail below the second scenario.

\subsubsection{Computing representative sequences from users' histories}
A recommendation made for a user is a generalization of behaviors observed for many other users of the system. To capture the short-term dynamics of users, we want to identify substrings of purchased items that characterize the possible short-term dynamics in the system. That is to say, we want to get sets of items ordered in time that the users of the system are likely to consider in the same order. Substrings that can account for user sequential dynamics are the ones that appear in many user's histories. We identify them by extracting frequent substrings \cite{GUSFIELD} that appear in at least \minCount{} user sequences and are at most of size \maxSize{}. \minCount{} and \maxSize{} are hyperparameters of our model. The obtained substrings constitute the set \freq{}. Three important points should be stressed here:
\begin{itemize}
	\item The set \freq{} is computed once at the beginning of \algo{} and then is used during the learning phase;
	\item The computation time of \freq{} takes at most ten seconds in \modif{our} experiments and is therefore not a computational bottleneck;
	\item Each substring of a frequent  substring is also frequent and thus belongs to \freq.
\end{itemize}

\subsubsection{Deriving from \freq{} a personalized sequence for $u$}
Once \freq{} \modif{is} computed, the objective is to find which substring to use as personalized context for user $u$ at \modif{position} $t$. This substring, denoted $\freqUserToTMinusOne$, must be included in the user's history while allowing the latter to contain additional items. To do that, we adapt the ``\emph{Exact matching with wildcards}'' algorithm
\cite{GUSFIELD} to compute the longest substring among the substrings in \freq{} that end to the most recent item in $\seqUserToTMinusOne$ that belongs to \freq{}. The function is presented in Algorithm~\ref{algo}.
It consists to pick up the most recent item of $\seqUserToTMinusOne$ that matches a substring of $\freq$. After that, it identifies the longest substring in $\freq$ that ends with the previously found item and which is contained in the user sequence. If this process ends without any substring matching, $\seqUser^{t-1}$ is taken as personalized context for $u$ at \modif{position} $t$, where $\seqUser^{t-1}$ is the most recent item considered by $u$ excluding $\groundTruth$, the ground truth item. It allows our model to always consider sequential dynamics. It is important to notice that the personalized context for a given user relies on its last actions and therefore varies according to \modif{position} $t$.

As an example, let us consider the set of frequent substrings $\freq = \{\langle 0 \rangle,\langle 1 \rangle,$ $ \langle 2 \rangle, \langle 3 \rangle, \langle 4 \rangle, \langle0,1\rangle, \langle 1,3 \rangle, \langle 0,1,3 \rangle, \langle 1,2 \rangle, \langle 2,4 \rangle, \langle 1,2,4 \rangle \}$ and consider the user sequence $s_{u}^{[1,7[} = \langle 0,1,2,3,4,5 \rangle$. The most recent item that can be exploited is $4$ as item $5$ does not appear in $\freq$. The longest sequence of \freq{} that ends with $4$ and that only contains items of $s_{u}^{[1,7[}$ in the same order is $ m_{s_{u}^{[1,7[}}=\langle 1,2,4\rangle$. Indeed, the substring $\langle 1,2,*,4 \rangle$ matches $s_{u}^{[2,5]}$ with $*$ a wildcard. If the user sequence is $s_{u}^{[1,4[}= \langle 7,8,9 \rangle$, none of the substrings of \freq{} matches $s_{u}^{[1,4[}$, and $m_{s_{u}^{[1,4[}}=\langle 9\rangle$.

\begin{algorithm}[htb]
	\caption{ExactMatchWithWildCard$(\seqUserToTMinusOne,\freq)$\label{algo}}
	\KwIn{A user sequence $\seqUserToTMinusOne = \langle\seqUser^{1},\cdots, \seqUser^{t-1}\rangle$ and the set of frequent sequences $\freq$ }
	\KwOut{ $\freqUserToTMinusOne$, the longest substring among the substrings in \freq{} that end to the most recent item in $\seqUserToTMinusOne$ that belongs to \freq{}, or $\seqUser^{t-1}$ if any.}
	$sequence \leftarrow  \langle\seqUser^{1},\cdots, \seqUser^{t-1}\rangle$\\ 
	$path \leftarrow \langle\rangle$\\
	\For{ $i=t-1$ \textbf{downto} 1}{	
		$item \leftarrow s_u^i$\\
		\If{$path = \langle\rangle$}{
			\If{$item \in \freq$}{
				$path\leftarrow \langle item\rangle$\\
			}
		}
		\Else{
			\If{$ \langle item\rangle\cdot path \in \freq$}{
				$path \leftarrow \langle item\rangle\cdot path$ \Comment{The concatenation of strings $ \langle item\rangle$ and $path$}\\
			}
		}
	}
	
	\If{$path = \langle\rangle$}{
		$path\leftarrow \langle \seqUser^{t-1}\rangle$\\
	}
	
	$return\ path $ \Comment{Here $\freqUserToTMinusOne \leftarrow path$ }\\
\end{algorithm}

\subsubsection{Constraining $P_i$ with short-term dynamics}
The personalized sequence $\freqUserToTMinusOne$ is then used to constrain the latent vector $P_i$ to be as close as possible to the items contained in $\freqUserToTMinusOne$. A damping factor $\eta_r$, depending on the item position $r$, is used to increase the importance of recent items of $\freqUserToTMinusOne$. With $P_{\RItemFreqUserToTMinusOne}$ be the vector corresponding to the item of  $\freqUserToTMinusOne$ at position $r$, we have:
\begin{eqnarray}
\label{eq2}
\predUserItemTime \propto -||\sum_{r=1}^{R} \eta_{r} P_{\RItemFreqUserToTMinusOne}-P_{i}||_{2}^{2}.
\end{eqnarray}
where $R=|\freqUserToTMinusOne|$ is the number of items in $\freqUserToTMinusOne$. The value of $\eta_{r}$ increases with $r$ to give more weight to recent items. To normalize the short-term part of \algo, we make $\eta_{r}$ follow the softmax (normalized exponential) \cite{Bishop:2006} of the following linear function:
\begin{eqnarray*}
	\eta_{r} = \frac{e^{\frac{r}{R}-1}}{\sum_{r=1}^{R} e^{\frac{r}{R}-1}}
\end{eqnarray*}
Let us consider the same example as above where $ m_{s_{u}^{[1,7[}}=\langle 1,2,4\rangle$ and $\langle 1,2,*,4 \rangle$ matches $s_{u}^{[2,5]}$. Positions of items $\langle     1,2,4\rangle$ are respectively 1, 2 and 3 and $|m_{s_{u}^{[1,7[}}| = 3$ because we do not take into consideration wildcard item $*$. Thus $\eta_{1}\approx0.23$, $\eta_{2}\approx0.321$ and $\eta_{3}\approx0.448$. As expected the most recent item $4$ has a greater importance compared to items $1$ and $2$.  

\subsection{The long-term and short-term metric embedding model}
The embedding of items into a metric space has two main advantages. \modif{First, it brings better generalization as Euclidean distances preserve the triangle inequalities}. Second, it makes it possible to fully unify the long and short-term dynamics (as expressed by equations \ref{eq1} and \ref{eq2}) of each item into a single embedding vector resulting from the computation of one distance:
\begin{eqnarray*}
	\predUserItemTime \propto -||(\text{Long-term} + \text{Short-term})-P_{i}||_{2}^{2}.
\end{eqnarray*}
It is also usual to add a bias term $\beta_i$ specific to each item, and a hyperparameter $\gamma$ to weigh the importance between long-term and short-term dynamics:
\begin{eqnarray}
\label{eq:tot1}
\predUserItemTime \propto -\Big(\beta_{i} &+& ||\big(\gamma \frac{1}{|\setUserXToTMinusOne \setminus \{i\}|^{\alpha}}  \sum_{j \in \setUserXToTMinusOne \setminus \{i\}} P_{j} \nonumber\\
&+& (1-\gamma) \sum_{r=1}^{R} \eta_{r}
P_{\RItemFreqUserToTMinusOne}\big)
-P_{i}||_{2}^{2}\Big)
\end{eqnarray}

\subsection{Bayesian Personalized Ranking optimization criterion} \label{sec:opt}
The goal for a sequential recommender system is, for all users, to rank the ground-truth item higher than all other items. Bayesian Personalized Ranking (BPR) \cite{BPR} formalizes this problem as maximizing the posterior probability of the model parameters $\theta$, given the order relation  $\TotalOrder$ on items: $p(\theta\mid \TotalOrder)$.
Using Bayes' rule, the probability is proportional to $p(\TotalOrder\mid \theta)p(\theta)$. The goal is thus to identify the parameters that maximize the likelihood of correctly ordering items. It is formally expressed as having $i \text{ } \TotalOrder{} \text{ } j$, which means that $i$ is ranked higher than item $j$ for user $u$ at \modif{position} $t$ with $i=\groundTruth$ the ground truth item. Assuming independence of \modif{items, their orders and users}, this leads to estimate the model parameters by the maximum a posteriori probability (MAP):

\begin{eqnarray}
\label{eq:optpb}
\underset{\theta}{\text{arg max}}
&& = \ln\prod_{u \in U}\prod^{|\seqUser|}_{t = 2}\prod_{j \neq \groundTruth} \text{ } p(\groundTruth \TotalOrder j|\theta) \text{ } p(\theta)\nonumber \\
&& = \sum_{u \in U}\sum^{|\seqUser|}_{t = 2}\sum_{j \neq \groundTruth} \ln p(\groundTruth \TotalOrder j|\theta) +\ln p(\theta)
\end{eqnarray}

The parameters of \algo{} model are the embedded vectors $P_i$ and the bias terms $\beta_i$ for $i\in I$. $p(\groundTruth \TotalOrder j|\theta)$ is the probability that the ground truth item $\groundTruth$ is correctly ranked with respect to $j$ by the model, that is:
\begin{align*}
p(\groundTruth \TotalOrder j\mid\theta) &=  p(\predUsertemTruthTime \TotalOrder \predUserItemJTime | \theta) \\
&=p(\predUsertemTruthTime - \predUserItemJTime \TotalOrder 0 | \theta)
\end{align*}
This quantity is approximated by $\sigma(\predUsertemTruthTime - \predUserItemJTime)$, where $\sigma(z) = \frac{1}{1+e^{-z}}$ is the logistic sigmoid function. Taking as prior probability for $\theta$ a normal distribution with zero mean and $\lambda_\theta I$ as variance-covariance matrix, the criterion to optimize (equation~\ref{eq:optpb}) becomes:
\begin{eqnarray}
\label{eq:OPt}
\underset{\theta}{\text{arg max}}
&& = \sum_{u \in U}\sum^{|\seqUser|}_{t = 2}\sum_{j \neq \groundTruth}  \ln\sigma(\predUsertemTruthTime  - \predUserItemJTime) -\lambda_\theta||\theta||^2,
\end{eqnarray}
where $\lambda_\theta$ is a regularization hyperparameter.

\subsection{Model training}
\algo{} learns the embedded vectors $P_i$ and the bias terms $\beta_i$ by maximizing equation~\ref{eq:OPt}. Hyperparameters -- that is to say $\alpha$, $\gamma$, $\maxSize$, $\minCount$, $\MaxLenghtUP$ and regularization hyperparameters $\lambda_\theta$ -- are chosen using a grid search strategy. The learning rate as well as $k$, the length $k$ of embedded vectors, are fixed by the analyst. The hyperparameter $k$ must be chosen knowing that the larger it is, the more precise the item vectors and the more costly the computation of the model.
  
Item embedding $P_i$ are randomly initialized using Xavier initialization \cite{xavier} and the bias terms $\beta_i$ are initialized to zero.

The parameters are learned using a variant of Stochastic Gradient Descent (SGD) called \textit{Adam} optimizer \cite{adam} with a batch size of 128 examples. For each batch, it consists in uniformly sampling a user $u$, a \modif{position} $t$, that gives the positive item $i=s_u^t$, and a 'negative' item $j$.

The regularization term is computed over all items in $I$, contrary to what is done in TransRec, PRME and FPMC which only regularize items and users present in the batch. In \algo, it is possible to do that as only items are embedded and thus global regularization can be done at small cost, while providing better results.

\subsection{Recommendation}
Once \algo{} has been trained, it can be used to make recommendations. Considering the past actions of a user $\seqUserToT$, the most appropriate frequent sequence $\freqUserToT$ is found and used in equation~\ref{eq:tot1} with $i$ a candidate item. The item with the highest $\predUserItemTime$ value (or equivalently the smallest Euclidean distance) is recommended. Note that \algo{} can recommend more than one item, e.g. the Top-N items with the $N$ smallest Euclidean distances.

\subsection{Discussion}

As said before, one of the key characteristics of \algo{} is to only embed items, like SASRec and FMC do. This characteristic brings two advantages compared to other models that embed items and users (like FPMC, PRME, TransRec and CASER). First, it has a smaller space complexity and second it is able to make recommendations for new users without suffering from the cold start problem.

\paragraph{Space Complexity:} The learned parameters of \algo{} are the item embeddings -- each being of dimension $k$ -- and the items bias terms. This results in a number of parameters in $O(|I|\times k + |I|)$. Thus, the complexity of our model does not grow with the number of users, unlike other models that embed users and items.

\paragraph{Recommending items to new users:} We can say that \algo{} does not suffer from the cold-start problem, since it can make recommendations to new users who have interacted only with a single item of the system. We show in the next section that \algo{} \modif{has good performance} compared to other models when it comes to the problem of cold start users.

\section{Experiments} \label{sec:xp}

In this section, we present a thorough empirical study. We first begin by describing the \modif{12} real-world datasets we consider, as well as the questions these experiments aim to answer. Then, we report an extensive comparison of \algo{} with \modif{9} state-of-the-art algorithms for sequential recommendation. Results demonstrate that  \algo{} outperforms state-of-the-algorithms according to several metrics in most of the cases. Finally, we provide a deeper analysis of \algo{}, especially how frequent substrings are actually used in sequential recommendation, the impact of the user preferences with regard to the sequential dynamics and how \algo{} can be tuned to give better results.  For reproducibility purposes, the source code and the data are made available~\footnote{\urlCode{}}.

\subsection{Datasets and aims}
To evaluate the performance of \algo{} on both sparse and dense datasets from different domains, we consider \modif{5} well-known benchmarks and introduce a new dataset:\\
\textbf{Amazon} was introduced by \cite{Image-based}. It contains Amazon product reviews from May 1996 to July 2014 from several product categories. We have chosen to use the 3 following diversified categories: {\em Automotive, Office product} and {\em video games}.
\\ \textbf{MovieLens 1M}\footnote{\url{http://grouplens.org/datasets/movielens/1m/}} \cite{Harper-ML} is a popular dataset including 1 million movie ratings from 6040 users between April 2000 and February 2003. We used the datasets pre-processed by selecting the most recent $x$ ratings for each user, $x \in \{5,10,20,30,50\}$. These datasets allow us to study the performance of \algo{} on the same dataset but with different levels on sparsity (i.e. MovieLens with the 5 most recent ratings will be more sparse than MovieLens with the 50 most recent ratings).
-\\ \textbf{Epinions} describes consumer reviews for the website Epinions from January 2001 to November 2013. This dataset was collected by the authors of \cite{DBLP:conf/cikm/ZhaoMK14}.
\\ \textbf{Foursquare} depicts a large number of user check-ins on the Foursquare website from December 2011 to April 2012.
\\ \modif{\textbf{Adressa} \cite{adressa} includes news articles (in Norwegian). The dataset was offered by Adresseavisen, a local newspaper company in Trondheim, Norway.}   
\\  \textbf{Visiativ}\footnote{\url{https://www.visiativ.com/en-us/}} is a new dataset that gathers the downloads of Computer-Aided-Design resources by engineers and designers from November 2014 to August 2018. This dataset seems relevant to us to evaluate temporal recommendation since the downloaded documents are tutorials that users read to train themselves and improve their comprehension of some specific software.

For each dataset, ratings are converted into implicit feedback and we only consider users and items that have at least 5 interactions. The main characteristics of these datasets are reported in Table~\ref{tab:datasets}.

\begin{table}[!htb]
	\begin{center}
		\caption{Main characteristics of the datasets.}
		\label{tab:datasets}
		\begin{tabular}{|p{0.2cm}|l|p{1cm}p{1cm}p{1.2cm}p{1.5cm}p{1.5cm}p{1cm}|}
		  \hline
		  & Datasets & \#Users& \#Items & \#Actions & \#A/\#U & \#A/\#I & Sparsity \\
		  \hline\hline
		   \multirow{4}{*}{\rotatebox{90}{Others}}
		    &Epinions   & 5015   & 8335    & 26932   & 5.37  & 3.23   & 99.94\% \\
		    &Foursquare & 43110  & 13335   & 306553  & 7.11  & 22.99  & 99.95\% \\
		       &Adressa    & 141933 & 3257    & 1861901 & 13.12 & 571.66 & 99.60\% \\
		    &Visiativ   & 1398   & 590     & 16417   & 11.74 & 27.83  & 98.01\% \\ \cline{1-8}
		   \multirow{3}{*}{\rotatebox{90}{Amazon}}
		    &Ama-Auto   & 34316  & 40287   & 183573  & 5.35  & 4.56   & 99.99\%  \\
		    &Ama-Office & 16716  & 22357   & 128070  & 7.66  & 5.73   & 99.97\%  \\
		    &Ama-Game   & 31013  & 23715   & 287107  & 9.26  & 12.11  & 99.96\% \\ \cline{1-8}
		   \multirow{5}{*}{\rotatebox{90}{MovieLens}}
		    &ML-5       & 6040   & 2848    & 30175   & 5.00  & 10.60  & 99.82\%  \\
		    &ML-10      & 6040   & 3114    & 59610   & 9.87  & 19.14  & 99.68\%  \\
		    &ML-20      & 6040   & 3324    & 111059  & 18.39 & 33.41  & 99.45\%  \\
		    &ML-30      & 6040   & 3391    & 152160  & 25.19 & 44.87  & 99.26\%  \\
		    &ML-50      & 6040   & 3467    & 215676  & 35.71 & 62.21  & 98.87\%  \\ \hline
		\end{tabular}
	\end{center}
\end{table}

To evaluate the performance and the limits of \algo, we propose to answer the following questions:
\begin{itemize}
	\item What are the performances of \algo{} compared to those of state-of-the-art algorithms for sequential recommendation for sparse and dense datasets? What about \algo's performance in presence of cold-start users?  
	{ \item Does \algo{} take benefit from sequential behaviors in a better way than Markov chains of fixed order do?
	\item What is the most important component? The user preferences? The sequential dynamics or both? 
	\item \modif{What about the recommendations? Does \algo{} provide diverse recommendations?}
	 }
\end{itemize}

\subsection{Comparison methods}
We compare \algo{} to \modif{9} state-of-the-art methods designed for both
item and sequential recommendation:
\begin{itemize}
	\item Popularity (\textbf{POP}), the naive baseline that ranks items according to their popularity \cite{BookSR};
	\item {Bayesian Personalized Ranking (\textbf{BPR})} \cite{BPR}, that recommends items by considering only user preferences with matrix factorization techniques;
	\item {Factorized Markov Chains (\textbf{FMC})} \cite{FPMC}, that is based on the factorization of the item-to-item transition matrix;
	\item {Factorized Personalized Markov Chains (\textbf{FPMC})} \cite{FPMC}, that considers both user preferences and user dynamics thanks to matrix factorization and first-order Markov chains;
	\item {Personalized Ranking Metric Embedding (\textbf{PRME})} \cite{PRME}, that embeds user preferences and user dynamics into two Euclidean distances;
	\item {Translation-based Recommendation (\textbf{TransRec})} \cite{TransRec}, that unifies user preferences and sequential dynamics with translations; 
	\item {Convolutional Sequence Embedding Recommendation (\textbf{CASER})} \cite{CASER}, a CNN-based method that captures both user preferences and user sequential dynamics;
	\item {Self-Attentive Sequential Recommendation (\textbf{SASRec})} \cite{SASRec}, a self-attention based model that captures both user preferences and user sequential dynamics ;	
	\item\modif{{Session-based KNN (\textbf{S-KNN})} \cite{GRU4REC2}, a nearest-neighbor-based approach designed for session-recommendation. In our experiments, a user’s sequence is assimilate to a session \footnote{\modif{Note that we did not make experiments with \textbf{GRU4Rec} as it has already been shown that it is outperformed by both \textbf{CASER} \cite{CASER} and \textbf{SASRec} \cite{SASRec}, and sometimes also by \textbf{S-KNN} \cite{sessionBased}.}}}.
\end{itemize}

Table~\ref{tab:models} provides a comparison of the above methods according to several criteria: whether they are personalized, sequentially aware, metric-based, integrated into a unified model, based on personalized order Markov chains or on a model that only embed items. Indications about the time needed to train the models are also reported in the last column. For fair comparisons, we return the time required for 25 epochs on Amazon Office. As we can see, the time needed for \algo{} is comparable to the one of other methods. It includes the time needed to extract frequent substrings.

\begin{table}[htb]
  \begin{center}
    \caption{Overview of the different models.  \textbf{P}: Personalized?, \textbf{S}: Sequentially-aware?, \textbf{M}: Metric-based?, \textbf{U}: Unified model?, \textbf{O}: personalized Order Markov chains?, \textbf{I}: embed only items?, \textbf{T}: Time (in seconds) for the Amazon Office dataset with 25 epochs.}
    \label{tab:models}
    \scalebox{1}{
      \begin{tabular}{|l|c|c|c|c|c|c|c|}
        \hline
		    {Property} & P & S & M & U & O & I & T \\ \hline\hline
		    Pop & \textcolor{red}{\xmark} & \textcolor{red}{\xmark} & \textcolor{red}{\xmark} & \textcolor{red}{\xmark} & \textcolor{red}{\xmark} & \textcolor{red}{\xmark} &  $\sim$1s\\ \hline
		    BPR& \textcolor{green}{\cmark} & \textcolor{red}{\xmark} & \textcolor{red}{\xmark} & \textcolor{red}{\xmark} & \textcolor{red}{\xmark} & \textcolor{red}{\xmark} & $\sim$21\\ \hline
		    FMC&\textcolor{red}{\xmark}  & \textcolor{green}{\cmark} & \textcolor{red}{\xmark} & \textcolor{red}{\xmark} & \textcolor{red}{\xmark} & \textcolor{green}{\cmark} & $\sim$19  \\ \hline
		    FPMC&	\textcolor{green}{\cmark}  & \textcolor{green}{\cmark} & \textcolor{red}{\xmark} & \textcolor{red}{\xmark} & \textcolor{red}{\xmark} & \textcolor{red}{\xmark} & $\sim$26s   \\ \hline
		    PRME&	 \textcolor{green}{\cmark}  & \textcolor{green}{\cmark} &  \textcolor{green}{\cmark} & \textcolor{red}{\xmark} & \textcolor{red}{\xmark} & \textcolor{red}{\xmark} & $\sim$25s \\ \hline
		    TransRec&   \textcolor{green}{\cmark}  & \textcolor{green}{\cmark} &  \textcolor{green}{\cmark} &  \textcolor{green}{\cmark} & \textcolor{red}{\xmark} & \textcolor{red}{\xmark} & $\sim$27s \\ \hline
		    \modif{S-KNN} & \textcolor{green}{\cmark}  & \textcolor{green}{\cmark} & \textcolor{red}{\xmark} &  \textcolor{red}{\xmark} & \textcolor{red}{\xmark} & \textcolor{red}{\xmark} & $\sim$1s \\ \hline
			CASER&   \textcolor{green}{\cmark}  & \textcolor{green}{\cmark} &  \textcolor{red}{\xmark} &  \textcolor{green}{\cmark} & \textcolor{red}{\xmark} & \textcolor{red}{\xmark} & $\sim$75s \\ \hline
			SASRec&   \textcolor{green}{\cmark}  & \textcolor{green}{\cmark} &  \textcolor{red}{\xmark} &  \textcolor{green}{\cmark} & \textcolor{red}{\xmark} & \textcolor{green}{\cmark} & $\sim$20s \\ \hline
			\algo{} &\textcolor{green}{\cmark}  & \textcolor{green}{\cmark} &  \textcolor{green}{\cmark} &  \textcolor{green}{\cmark} & \textcolor{green}{\cmark} & \textcolor{green}{\cmark} & $\sim$30s  \\ \hline
		\end{tabular}
		}
	\end{center}
\end{table}

\subsection{Experimental settings}
\modif{We apply the same partition as \cite{FrameworkLearningImplictFeedback,NeuMF,TransRec,SASRec} for user sequences (also known as \textit{leave-one-out} evaluation). Indeed, for each dataset, the user sequences are split into 3 parts:}
\begin{enumerate}
\item The most recent item is used for the test. It is named \textit{ground-truth item} and denoted $\groundTruthBis$;
\item The second most recent item is used for the validation when learning the model;
\item Other items of the user sequence are used to train the models.
\end{enumerate}
The performances of the models are assessed by three widely used metrics for sequential recommendation \cite{BPR,TransRec,SASRec}: \\

\ptitle{Area Under the ROC Curve (AUC)}
\begin{eqnarray*}
	AUC = \frac{1}{|U|} \sum_{u \in U} \frac{1}{|I \setminus I_{s_u}|} \sum_{j \in I \setminus I_{s_u}} \mathbbm{1} (\widehat{p}_{u,g_u,t} > \widehat{p}_{u,j,t}),
\end{eqnarray*}
where the indicator function $\mathbbm{1}(b)$ returns 1 if its argument \textit{b} is \textit{True}, \textit{0} otherwise. This measure calculates how high the ground-truth item of each user has been ranked in average.\\

\ptitle{Hit Rate at position X (HIT\_X)}
\begin{eqnarray*}
	HIT\_X = \frac{1}{|U|} \sum_{u \in U} \mathbbm{1}(\RankingGroundTruthBis \leq R_X),
\end{eqnarray*}
where $\RankingGroundTruthBis$ is the ranking of the ground-truth item and $R_X$ is the Xth ranking. The HIT\_X function returns the average number of times the ground-truth item is ranked in the top X items. We compute HIT\_5, HIT\_10, HIT\_25 and HIT\_50. Note that HIT\_X is equal to Recall@X and proportional to Precision@X, two others common metrics in recommendation, as we only have one test item per user.\\

\ptitle{Normalized Discounted Cumulative Gain at position X (NDCG\_X)}
\begin{eqnarray*}
	NDCG\_X = \frac{1}{|U|} \sum_{u \in U} \frac{\mathbbm{1}(\RankingGroundTruthBis \leq R_X)}{log_{2}(\RankingGroundTruthBis+1)}.
\end{eqnarray*}
The NDCG\_X is a position-aware metric which assigns larger weights to higher positions.   We compute NDCG\_5, NDCG\_10, NDCG\_25 and NDCG\_50.

\modif{

We also consider two metrics to assess how diverse are the recommendations made by a model. 
	
\ptitle{Popularity rate X (Pop\_X)}
The POP\_X is the proportion of predicted items that are similar to the X most popular items. We can note that the POP model (the most popular) should have a Pop\_X value equal to $1$. However, it is not always the case since already consumed items are not recommended to user. 

\ptitle{Diversity rate X (Div\_X)} 
The DIV\_X is the proportion of items that are predicted at least once to all users (the number of unique predicted items divided by the total number of items). Div\_X is sometimes referred to as aggregate diversity \cite{DivMetric}.
}

Contrary to \cite{SASRec,NeuMF,KSR}, which follow the strategy of taking a sample of $x$ negative items during the evaluation (i.e. items that a user has not interacted with), we compute the metrics with all possible negative items \modif{(also known as \textit{TrainItems} \cite{comparativeRS}) to have an impartial assessment of all methods with exact measures instead of approximated ones}. 

\algo{} is implemented using \textit{TensorFlow} and the \textit{Adam} optimizer \cite{adam}. In order to make a fair comparison, we implement BPR, FMC, FPMC, PRME, TransRec with the same architecture as \algo. For CASER and SASRec, we use the code provided by the authors. To limit the numbers of combinations to explore with grid search, we fix the following parameters for all models: The learning rate to $0.001$, the batch size to 128, the dimension of the learned latent vectors $k$ is set to 10 and we stop the training if there is no improvement of the AUC validation for 250 epochs. All regularization hyperparameters are taken in $\{0,0.001,0.01,0.1,1\}$. For models with other hyperparameters we have tried those given by the authors, except for CASER which requires too many hyperparameters: There are about 50000 possible combinations. Regarding specific hyperparameters of \algo{}, $\alpha$ is taken in $\{0.3, 0.5, 0.7, 1.0\}$, \modif{$\gamma$ in $\{0.3, 0.5, 0.7\}$} \footnote{\modif{Only for Foursquare dataset, we observed that it is better not to have $\gamma$ (remove $\gamma$ and $(1-\gamma)$ in Equation~\ref{eq:tot1}). It is noteworthy that recommendations may be different to the case where $\gamma=0.5$ due to the bias terms $\beta_i$.}}, $\minCount$ has for default value $2$, and $\maxSize$ takes its value between $3$ and $5$ for small datasets (Visiativ, Epinions, ML-5) and between $8$ and $15$ for other datasets\footnote{Best hyperparameters for each dataset are reported in supplementary material \cite{REBUS_supplementary}.}.

\subsection{Performance study}\label{sec:perf_study}
The performances of the different methods on every dataset are reported in Table~\ref{tab:results} and Table~\ref{tab:resultsbis}\footnote{We only show HIT\_25, HIT\_50, NDCG\_25 and NDCG\_50 in the tables. HIT\_5, HIT\_10, NDCG\_5 and NDCG\_10 are reported in supplementary material.}. We can observe that \algo{} outperforms other models on most of the datasets, having the best AUC value. \algo{} also performs  well on  all other metrics, with an average rank between \modif{2.83 to 3.75, an improvement  between 1.35\% to 6.94\%} over the best competitor, and the best HIT\_25, HIT\_50, NDCG\_25 and NDCG\_50 in overall. For HIT\_5, HIT\_10, NDCG\_5 and NDCG\_10, we respectively found improvements of \modif{7.15\%, 3.22\%, 11.56\% and 8.74\%} compared to the best competitor. However, these high percentages have to be nuanced because the obtained values are often low (i.e., in comparison to AUC)  and the results vary a lot from one dataset to another. 

\modif{
In general, BPR-MF outperforms FMC. This confirms that user preferences have a more important role than the sequential dynamics in the recommendations of the next-item. As expected, SASRec performs very well on dense datasets (i.e., ML30, ML50 where $\#A/\#U > 20$). However, this model is outperformed by several others on sparse datasets. We can observe that \algo{} also outperforms PRME and FPMC on sparse datasets. It gives evidence that having independent latent vectors to model user preferences and sequential dynamics is not an advantage for sparse datasets.
S-KNN exhibits bad AUC performance because of its architecture \footnote{\modif{Performances of others nearest-neighbor-based approaches as Item-based KNN and Sequence-Aware Extensions (V-S-KNN, S-S-KNN and SF-S-KNN) are reported in supplementary material.}}: S-KNN cannot rank items that are not in the neighborhood of the user target. Besides, S-KNN reports fair performance on HIT\_X and NDCG\_X and outperforms all models on Amazon datasets -- this confirms the observations made by the authors of  \cite{sessionBased} -- but it is outperformed by \algo{} for all other datasets. 

The improvement of \algo{} according to each model is reported in Table~\ref{tab:improvement}. Considering all datasets, \algo{} outperforms all the studied models. However, some models exhibits better performances when only focusing on MovieLens datasets (e.g., SASRec, TransRec, FPMC and FMC for which \algo{} has negative improvements).

}

\begin{table}[!h]
	\centering
	\caption{\modif{AUC, HIT\_25, HIT\_50, NDCG\_25 and NDCG\_50 for the different models on Epinions, Foursquare, Adressa, Visiativ, Amazon-Automotive, Amazon-Office-Product and Amazon-Video-Games datasets.
		The last row, called \textit{Improv. vs Best}, shows the improvement in percentage of our method compared to the other best model (best obtained results are in bold and best obtained results for concurrent models are underlined).
		The last column is the average performance on these 7 datasets, the average of all datasets is included in Table~\ref{tab:resultsbis}.}}
	\label{tab:results}
	\scalebox{0.77}{
		\rotatebox{0}{
			\begin{tabular}{c|c|ccccccc|c}
				& \textbf{Metric} & \textbf{Epinions} & \textbf{Foursq} & \textbf{Adressa} & \textbf{Visiativ} & \textbf{Auto} & \textbf{Office} & \textbf{Games} & \textbf{Avg} \\ \hline
				
			POP & AUC & 0.4575 & 0.9169 & 0.9582 & 0.7864 & 0.5856 & 0.6412 & 0.7484 & 0.7277 \\
			& HIT25 & 2.25\% & 46.66\% & 19.39\% & 32.04\% & 2.54\% & 0.62\% & 3.50\% & 15.29\% \\
			& HIT50 & 3.42\% & 55.65\% & 30.89\% & 48.24\% & 3.75\% & 1.67\% & 5.16\% & 21.25\% \\
			& NDGC25 & 0.80\% & 18.92\% & 7.21\% & 12.87\% & 0.96\% & 0.15\% & 1.37\% & 6.04\% \\
			& NDGC50 & 1.03\% & 20.68\% & 9.40\% & 15.97\% & 1.19\% & 0.36\% & 1.69\% & 7.19\% \\
			FMC & AUC & 0.5421 & 0.9508 & 0.9842 & 0.8354 & 0.6251 & 0.6771 & 0.8473 & 0.7803 \\
			& HIT25 & 1.67\% & 53.06\% & 64.23\% & 45.66\% & 2.57\% & 1.35\% & 10.64\% & 25.60\% \\
			& HIT50 & 2.63\% & 64.19\% & 77.99\% & 58.78\% & 3.78\% & 2.67\% & 15.71\% & 32.25\% \\
			& NDGC25 & 0.77\% & 26.13\% & 29.19\% & 23.24\% & 1.05\% & 0.52\% & 4.19\% & 12.15\% \\
			& NDGC50 & 0.95\% & 28.29\% & 31.84\% & 25.75\% & 1.28\% & 0.77\% & 5.16\% & 13.44\% \\
			BPR & AUC & 0.5593 & 0.9474 & 0.9656 & 0.8270 & 0.6649 & 0.7072 & 0.8590 & 0.7901 \\
			& HIT25 & \underline{2.88\%} & 46.95\% & 25.83\% & 42.72\% & 2.90\% & 1.58\% & 7.82\% & 18.67\% \\
			& HIT50 & \underline{4.49\%} & 58.38\% & 40.16\% & 57.42\% & 4.59\% & 2.59\% & 12.42\% & 25.72\% \\
			& NDGC25 & 1.20\% & 20.44\% & 9.35\% & 17.13\% & 1.10\% & 0.63\% & 2.85\% & 7.53\% \\
			& NDGC50 & 1.51\% & 22.66\% & 12.10\% & 19.95\% & 1.43\% & 0.82\% & 3.73\% & 8.88\% \\
			FPMC & AUC & 0.5536 & 0.9492 & 0.9848 & 0.8433 & 0.6482 & 0.6958 & 0.8777 & 0.7932 \\
			& HIT25 & 1.63\% & 56.46\% & 65.36\% & 46.67\% & 2.36\% & 1.60\% & 12.28\% & 26.62\% \\
			& HIT50 & 2.51\% & 65.38\% & 79.39\% & 60.36\% & 3.64\% & 2.53\% & 18.33\% & 33.16\% \\
			& NDGC25 & 0.70\% & 30.88\% & 30.14\% & 22.68\% & 0.88\% & 0.56\% & 4.71\% & 12.94\% \\
			& NDGC50 & 0.87\% & 32.61\% & 32.85\% & 25.35\% & 1.13\% & 0.74\% & 5.87\% & 14.20\% \\
			PRME & AUC & 0.6071 & 0.9538 & 0.9849 & 0.8572 & 0.6749 & 0.7154 & 0.8759 & 0.8099 \\
			& HIT25 & 1.88\% & 55.64\% & 64.78\% & 45.09\% & 2.34\% & 2.78\% & 12.63\% & 26.45\% \\
			& HIT50 & 2.72\% & 67.23\% & 78.19\% & 59.43\% & 3.85\% & 4.87\% & 18.74\% & 33.58\% \\
			& NDGC25 & 0.85\% & 30.71\% & 30.33\% & 24.86\% & 0.85\% & 0.95\% & 4.89\% & 13.35\% \\
			& NDGC50 & 1.01\% & 32.97\% & 32.92\% & 27.61\% & 1.14\% & 1.34\% & 6.07\% & 14.72\% \\
			TransRec & AUC & 0.6138 & \underline{0.9619} & 0.9852 & 0.8647 & \underline{0.6991} & 0.7320 &\underline{ 0.8869} & 0.8205 \\
			& HIT25 & 1.98\% & 58.67\% & 65.70\% & 51.25\% & 3.98\% & 3.97\% & 10.76\% & \underline{28.04\%} \\
			& HIT50 & 3.14\% & 67.81\% & 78.81\% & 64.73\% & 5.99\% & 6.37\% & 16.51\% & \underline{34.77\%} \\
			& NDGC25 & 0.94\% & 33.13\% & \underline{31.99\% }& \underline{\textbf{26.66\%}} & 1.56\% & 1.38\% & 4.08\% & \underline{14.25\%} \\
			& NDGC50 & 1.16\% & 34.90\% & \underline{34.52\% }& \underline{\textbf{29.26\%}} & 1.95\% & 1.84\% & 5.18\% & \underline{15.55\%} \\
			S-KNN & AUC & 0.0684 & 0.8330 & 0.9367 & 0.8369 & 0.1476 & 0.2938 & 0.6322 & 0.5355 \\
			& HIT25 & 2.51\% & \underline{\textbf{63.90\%}} & 27.77\% & 44.87\% & \underline{\textbf{5.15\%}} & \underline{\textbf{5.36\%} }& \underline{\textbf{13.94\%}} & 23.36\% \\
			& HIT50 & 3.70\% & \underline{\textbf{70.75\%}} & 40.33\% & 61.15\% & \underline{\textbf{6.71\%}} & \underline{\textbf{7.32\%}} & \underline{1\textbf{9.54\%}} & 29.93\% \\
			& NDGC25 &\underline{ \textbf{1.32\%}} & \underline{45.92\%} & 12.97\% & 21.83\% & \underline{\textbf{2.49\%} }& \underline{\textbf{2.77\%}} & \underline{\textbf{5.82\%}} & 13.30\% \\
			& NDGC50 &\underline{ 1.55\% }& \underline{47.25\%} & 15.36\% & 24.95\% & \underline{\textbf{2.79\% }}& \underline{\textbf{3.15\%}} & \underline{\textbf{6.89\%}} & 14.56\% \\
			CASER & AUC & 0.6238 & 0.9259 & 0.9750 & 0.8544 & 0.6872 & \underline{\textbf{0.7510}} & 0.8282 & 0.8065 \\
			& HIT25 & 2.44\% & 46.45\% & 60.65\% & 46.74\% & 2.63\% & 1.42\% & 6.45\% & 23.83\% \\
			& HIT50 & 3.88\% & 56.73\% & 74.86\% & 61.08\% & 3.64\% & 2.50\% & 10.23\% & 30.42\% \\
			& NDGC25 & 0.96\% & 18.39\% & 26.29\% & 21.35\% & 0.94\% & 0.45\% & 2.37\% & 10.11\% \\
			& NDGC50 & 1.24\% & 20.40\% & 29.04\% & 24.12\% & 1.14\% & 0.66\% & 3.10\% & 11.38\% \\
			SASRec & AUC & \underline{0.6251} & 0.9617 & \underline{\textbf{0.9870}} & \underline{0.8731} & 0.6861 & 0.7392 & 0.8812 & \underline{0.8219} \\
			& HIT25 & 2.60\% & 55.04\% & \underline{\textbf{67.30\%} }& \underline{51.76\%} & 2.55\% & 3.65\% & 9.16\% & 27.44\% \\
			& HIT50 & 4.00\% & 64.97\% & \underline{\textbf{82.15\%}} & \underline{65.38\%} & 4.07\% & 6.12\% & 14.54\% & 34.46\% \\
			& NDGC25 & 1.03\% & 30.70\% & 29.23\% & 25.00\% & 0.95\% & 1.30\% & 3.56\% & 13.11\% \\
			& NDGC50 & 1.30\% & 32.62\% & 32.10\% & 27.61\% & 1.24\% & 1.77\% & 4.59\% & 14.46\% \\ \hline
			
			REBUS & AUC & \textbf{0.6524} & \textbf{0.9677} & 0.9854 & \textbf{0.8735} & \textbf{0.7184} & 0.7507 & \textbf{0.8934} & \textbf{0.8345} \\
			& HIT25 & \textbf{3.39\%} & 62.77\% & 66.73\% & \textbf{53.33\%} & 4.24\% & 4.12\% & 9.30\% &\textbf{ 29.13\%} \\
			& HIT50 & \textbf{5.32\%} & 70.41\% & 79.77\% & \textbf{67.17\%} & 6.20\% & 6.24\% & 14.61\% & \textbf{35.68\%} \\
			& NDGC25 & 1.31\% & \textbf{46.02\%} & \textbf{32.00\%} & 24.20\% & 1.68\% & 1.52\% & 3.52\% & \textbf{15.75\%} \\
			& NDGC50 & \textbf{1.68\%} & \textbf{47.49\%} & \textbf{34.52\%} & 26.86\% & 2.05\% & 1.92\% & 4.54\% & \textbf{17.01\%} \\
			
			Rank &  AUC  & 1 & 1 & 2 & 1 & 1 & 2 & 1 & 1.29 \\
			    &  HIT25 & 1 & 2 & 2 & 1 & 2 & 2 & 3 & 1.86 \\
		    	&  HIT50 & 1 & 2 & 2 & 1 & 2 & 3 & 6 & 2.43 \\
			   &  NDGC25 & 2 & 1 & 1 & 4 & 2 & 2 & 7 & 2.71 \\
			   &  NDGC50 & 1 & 1 & 1 & 4 & 2 & 2 & 7 & 2.57 \\ \hline
			
			Improv. & AUC    & 4.37\%  & 0.60\%  & -0.16\% & 0.05\%  & 2.76\%   & -0.04\%  & 0.73\%   & 1.53\% \\
			vs Best & HIT25  & 17.71\% & -1.77\% & -0.85\% & 3.03\%  & -17.67\% & -23.13\% & -33.29\% & 3.89\% \\
			        & HIT50  & 18.49\% & -0.48\% & -2.90\% & 2.74\%  & -7.60\%  & -14.75\% & -25.23\% & 2.62\% \\
		         	& NDGC25 & -0.76\% & 0.22\%  & 0.03\%  & -9.23\% & -32.53\% & -45.13\% & -39.52\% & 10.53\% \\
		         	& NDGC50 & 8.39\%  & 0.51\%  & 0.00\%  & -8.20\% & -26.52\% & -39.05\% & -34.11\% & 9.39\% \\ \hline
			
			\end{tabular}
		}
	}
\end{table}

\begin{table}[!h]
	\centering
	\caption{\modif{AUC, HIT\_25, HIT\_50, NDCG\_25 and NDCG\_50 for the different models on ML-5, ML-10, ML-20, ML-30 and ML-50 datasets.
		The last row, called \textit{Improv. vs Best}, shows the improvement in percentage of our method compared to the other best model (best obtained results are in bold and best obtained results for concurrent models are underlined).
		The last column is the average performance on the 12 datasets, including datasets of Table~\ref{tab:results}.}}
	\label{tab:resultsbis}
	\scalebox{0.80}{
		\rotatebox{0}{
			\begin{tabular}{c|c|cccccc|c}

			& \textbf{ Metric} & \textbf{ ML-5} & \textbf{ ML-10} & \textbf{ ML-20} & \textbf{ ML-30} & \textbf{ ML-50} & \textbf{ Avg(ML)} & \textbf{ Avg(All)} \\ \hline
				
			POP & AUC & 0.7352 & 0.7722 & 0.7919 & 0.7981 & 0.8032 & 0.7801 & 0.7496 \\
			& HIT25 & 7.48\% & 7.60\% & 7.57\% & 7.57\% & 7.72\% & 7.59\% & 12.08\% \\
			& HIT50 & 12.65\% & 12.93\% & 12.82\% & 12.72\% & 13.13\% & 12.85\% & 17.75\% \\
			& NDGC25 & 2.64\% & 2.75\% & 2.73\% & 2.66\% & 2.70\% & 2.70\% & 4.65\% \\
			& NDGC50 & 3.63\% & 3.76\% & 3.74\% & 3.64\% & 3.74\% & 3.70\% & 5.74\% \\
			FMC & AUC & 0.7414 & 0.8054 & 0.8446 & 0.8560 & 0.8689 & 0.8233 & 0.7982 \\
			& HIT25 & 9.84\% & 13.84\% & 16.44\% & 16.74\% & 18.63\% & 15.10\% & 21.22\% \\
			& HIT50 & 15.25\% & 20.60\% & 24.95\% & 25.58\% & 27.72\% & 22.82\% & 28.32\% \\
			& NDGC25 & 3.57\% & 5.72\% & 6.62\% & 6.69\% & 7.36\% & 5.99\% & 9.59\% \\
			& NDGC50 & 4.61\% & 7.01\% & 8.25\% & 8.39\% & 9.11\% & 7.47\% & 10.95\% \\
			BPR & AUC & 0.7623 & 0.8241 & 0.8517 & 0.8570 & 0.8629 & 0.8316 & 0.8074 \\
			& HIT25 & 9.65\% & 11.96\% & 11.49\% & 10.66\% & 11.24\% & 11.00\% & 15.47\% \\
			& HIT50 & 15.98\% & 19.74\% & 19.56\% & 18.74\% & 18.56\% & 18.52\% & 22.72\% \\
			& NDGC25 & 3.52\% & 4.16\% & 4.06\% & 3.68\% & 3.87\% & 3.86\% & 6.00\% \\
			& NDGC50 & 4.73\% & 5.65\% & 5.61\% & 5.22\% & 5.27\% & 5.30\% & 7.39\% \\
			FPMC & AUC & 0.7309 & 0.8150 & 0.8629 & 0.8766 & 0.8891 & 0.8349 & 0.8106 \\
			& HIT25 & 7.82\% & 14.65\% & \underline{\textbf{17.42\%}} & \underline{\textbf{18.65\%}} & \underline{\textbf{18.96\%}} & 15.50\% & 21.99\% \\
			& HIT50 & 13.26\% & 22.90\% & 26.28\% & 27.60\% & 28.63\% & 23.74\% & 29.23\% \\
			& NDGC25 & 2.87\% & 5.57\% & \underline{\textbf{6.63\%}} & \underline{\textbf{7.27\%}} & \underline{\textbf{7.41\%}} & 5.95\% & 10.02\% \\
			& NDGC50 & 3.91\% & 7.15\% & \underline{\textbf{8.33\%}} & \underline{\textbf{8.99\%}} & \underline{\textbf{9.26\%}} & 7.53\% & 11.42\% \\
			PRME & AUC & 0.7771 & 0.8302 & 0.8645 & 0.8765 & 0.8867 & 0.8470 & 0.8254 \\
			& HIT25 & 10.51\% & 15.58\% & 14.56\% & 16.19\% & 18.00\% & 14.97\% & 21.67\% \\
			& HIT50 & 16.94\% & 23.81\% & 24.39\% & 25.55\% & 27.14\% & 23.57\% & 29.41\% \\
			& NDGC25 & 3.78\% & 5.79\% & 5.16\% & 5.94\% & 6.62\% & 5.46\% & 10.06\% \\
			& NDGC50 & 5.01\% & 7.37\% & 7.04\% & 7.72\% & 8.37\% & 7.10\% & 11.55\% \\
			TransRec & AUC & 0.7900 & 0.8477 & 0.8736 & 0.8816 & 0.8883 & 0.8563 & 0.8354 \\
			& HIT25 & \underline{14.36\%} & \underline{\textbf{16.99\%}} & 16.49\% & 16.00\% & 17.09\% & 16.18\% & \underline{23.10\%} \\
			& HIT50 & \underline{22.45\%} & \underline{\textbf{25.60\%}} & \underline{\textbf{26.51\%}} & 26.30\% & 27.87\% & 25.75\% & \underline{31.01\%} \\
			& NDGC25 & \underline{\textbf{5.33\%}} & \underline{\textbf{6.30\%}} & 5.96\% & 5.95\% & 6.35\% & 5.98\% & \underline{10.80\%} \\
			& NDGC50 & \underline{6.88\%} & \underline{\textbf{7.95\%}} & 7.89\% & 7.93\% & 8.42\% & 7.81\% & \underline{12.32\%} \\
			S-KNN & AUC & 0.2778 & 0.7164 & 0.8115 & 0.8297 & 0.8421 & 0.6955 & 0.6022 \\
			& HIT25 & 9.46\% & 14.99\% & 12.85\% & 12.62\% & 11.97\% & 12.38\% & 18.78\% \\
			& HIT50 & 13.08\% & 22.62\% & 21.13\% & 20.50\% & 19.27\% & 19.32\% & 25.51\% \\
			& NDGC25 & 3.87\% & 5.55\% & 4.54\% & 4.48\% & 4.26\% & 4.54\% & 9.65\% \\
			& NDGC50 & 4.57\% & 7.01\% & 6.13\% & 5.99\% & 5.66\% & 5.87\% & 10.94\% \\
			CASER & AUC & 0.7493 & 0.8118 & 0.8534 & 0.8650 & 0.8764 & 0.8312 & 0.8168 \\
			& HIT25 & 7.40\% & 11.36\% & 14.27\% & 15.95\% & 18.66\% & 13.53\% & 19.54\% \\
			& HIT50 & 13.58\% & 18.20\% & 22.62\% & 24.57\% & 28.18\% & 21.43\% & 26.67\% \\
			& NDGC25 & 2.63\% & 4.06\% & 5.35\% & 5.89\% & 7.07\% & 5.00\% & 7.98\% \\
			& NDGC50 & 3.82\% & 5.37\% & 6.95\% & 7.54\% & 8.90\% & 6.52\% & 9.36\% \\
			SASRec & AUC & \underline{0.8000} & \underline{0.8537} & \underline{0.8760} & \underline{\textbf{0.8860}} & \underline{\textbf{0.8957}} & \underline{\textbf{0.8623}} & \underline{0.8387} \\
			& HIT25 & 12.77\% & 15.91\% & 16.11\% & 18.26\% & 18.55\% & \underline{\textbf{16.32\%}} & 22.80\% \\
			& HIT50 & 20.83\% & 25.32\% & 25.96\% & \underline{\textbf{28.20\%}} & \underline{\textbf{29.29\%}} & \underline{\textbf{25.92\%}} & 30.90\% \\
			& NDGC25 & 4.53\% & 6.08\% & 5.81\% & 6.97\% & 7.09\% & \underline{\textbf{6.10\%}} & 10.19\% \\
			& NDGC50 & 6.08\% & 7.89\% & 7.70\% & 8.87\% & 9.15\% & \underline{\textbf{7.94\%}} & 11.74\% \\ \hline
			
			REBUS & AUC & \textbf{0.8031} & \textbf{0.8563} & \textbf{0.8772} & 0.8826 & 0.8884 & 0.8615 &\textbf{ 0.8458} \\
			& HIT25 & \textbf{14.42\%} & 15.27\% & 15.60\% & 15.80\% & 16.72\% & 15.56\% & \textbf{23.48\%} \\
			& HIT50 & \textbf{22.79\%} & 25.50\% & 26.01\% & 26.21\% & 26.91\% & 25.48\% & \textbf{31.43\%} \\
			& NDGC25 & 5.32\% & 5.59\% & 5.60\% & 5.80\% & 6.03\% & 5.66\% & \textbf{11.55\%} \\
			& NDGC50 & \textbf{6.92\%} & 7.55\% & 7.59\% & 7.80\% & 7.98\% & 7.57\% & \textbf{13.07\%} \\
			
			Rank &  AUC & 1 & 1 & 1 & 2 & 3 & 1.60 & 1.42 \\
			&  HIT25 & 1 & 4 & 5 & 7 & 7 & 4.80 & 3.08 \\
			&  HIT50 & 1 & 2 & 3 & 4 & 7 & 3.40 & 2.83 \\
			&  NDGC25 & 2 & 5 & 5 & 7 & 7 & 5.20 & 3.75 \\
			&  NDGC50 & 1 & 3 & 5 & 5 & 7 & 4.20 & 3.25 \\ \hline
			
			Improv. & AUC   & 0.39\% & 0.30\% & 0.14\% & -0.38\% & -0.82\% & -0.09\% &\textbf{ 0.85\%} \\
			vs Best & HIT25 & 0.42\% & -10.12\% & -10.45\% & -15.28\% & -11.81\% & -4.66\% & \textbf{1.65\%} \\
				    & HIT50 & 1.51\% & -0.39\% & -1.89\% & -7.06\% & -8.13\% & -1.70\% & \textbf{1.35\%} \\
			     	& NDGC25 & -0.19\% & -11.27\% & -15.54\% & -20.22\% & -18.62\% & -7.21\% & \textbf{6.94\%} \\
					& NDGC50 & 0.58\% & -5.03\% & -8.88\% & -13.24\% & -13.82\% & -4.66\% & \textbf{6.09\%} \\ \hline
				
			\end{tabular}
		}
	}
\end{table}

\begin{table}[htb]
	\centering
	\caption{\modif{Average improvement of \algo{} compared to each model on  others datasets (e.g., Epinions, Foursquare, Adressa, Visiativ, Amazon-Automotive, Amazon-Office-Product and Amazon-Video-Games datasets), ML datasets (e.g., ML-5, ML-10, ML-20, ML-30 and ML-50) datasets) and all datasets.}}
	\label{tab:improvement}
	\scalebox{0.77}{
		\rotatebox{0}{
			\begin{tabular}{c|c|ccccccccc}
			{\rotatebox{90}{}}	& {\rotatebox{90}{\textbf{ Metric}}} & {\rotatebox{90}{\textbf{SASRec}}} & {\rotatebox{90}{\textbf{TransRec}}} & {\rotatebox{90}{\textbf{PRME}}} & {\rotatebox{90}{\textbf{CASER}}} & {\rotatebox{90}{\textbf{FPMC}}} & {\rotatebox{90}{\textbf{BPR}}} & {\rotatebox{90}{\textbf{FMC}}} & {\rotatebox{90}{\textbf{POP}}} & {\rotatebox{90}{\textbf{S-KNN}}} \\ \hline
				
				& AUC & 1.53\% & 1.71\% & 3.04\% & 3.47\% & 5.21\% & 5.62\% & 6.95\% & 14.68\% & 55.84\% \\
				& HIT25 & 6.16\% & 3.89\% & 10.13\% & 22.24\% & 9.43\% & 56.03\% & 13.79\% & 90.52\% & 24.70\% \\
				Avg& HIT50 & 3.54\% & 2.62\% & 6.25\% & 17.29\% & 7.60\% & 38.72\% & 10.64\% & 67.91\% & 19.21\% \\
				Others& NDGC25 & 20.14\% & 10.53\% & 17.98\% & 55.79\% & 21.72\% & 109.16\% & 29.63\% & 160.76\% & 18.42\% \\
				& NDGC50 & 17.63\% & 9.39\% & 15.56\% & 49.47\% & 19.79\% & 91.55\% & 26.56\% & 136.58\% & 16.83\% \\ \hline
				
				& AUC & -0.09\% & 0.61\% & 1.71\% & 3.65\% & 3.19\% & 3.60\% & 4.64\% & 10.43\% & 23.87\% \\
				& HIT25 & -4.66\% & -3.83\% & 3.94\% & 15.00\% & 0.39\% & 41.45\% & 3.05\% & 105.01\% & 25.69\% \\
				Avg& HIT50 & -1.70\% & -1.05\% & 8.10\% & 18.90\% & 7.33\% & 37.58\% & 11.66\% & 98.29\% & 31.88\% \\
				ML& NDGC25 & -7.21\% & -5.35\% & 3.66\% & 13.20\% & -4.87\% & 46.63\% & -5.51\% & 109.63\% & 24.67\% \\
				& NDGC50 & -4.66\% & -3.07\% & 6.62\% & 16.10\% & 0.53\% & 42.83\% & 1.34\% & 104.59\% & 28.96\% \\ \hline
				
				& AUC & 0.85\% & 1.24\% & 2.47\% & 3.55\% & 4.34\% & 4.76\% & 5.96\% & 12.83\% & 40.45\% \\
				& HIT25 & 2.98\% & 1.65\% & 8.35\% & 20.16\% & 6.78\% & 51.78\% & 10.65\% & 94.37\% & 25.03\% \\ 
				Avg& HIT50 & 1.72\% & 1.35\% & 6.87\% & 17.85\% & 7.53\% & 38.34\% & 10.98\% & 77.07\% & 23.21\% \\
				All& NDGC25 & 13.35\% & 6.94\% & 14.81\% & 44.74\% & 15.27\% & 92.50\% & 20.44\% & 148.39\% & 19.69\% \\
				& NDGC50 & 11.33\% & 6.09\% & 13.16\% & 39.64\% & 14.45\% & 76.86\% & 19.36\% & 127.70\% & 19.47\% \\ \hline
				
			\end{tabular}
		}
	}
\end{table}

\begin{figure}[htb]
	\begin{center}
  		\includegraphics[width=1\columnwidth]{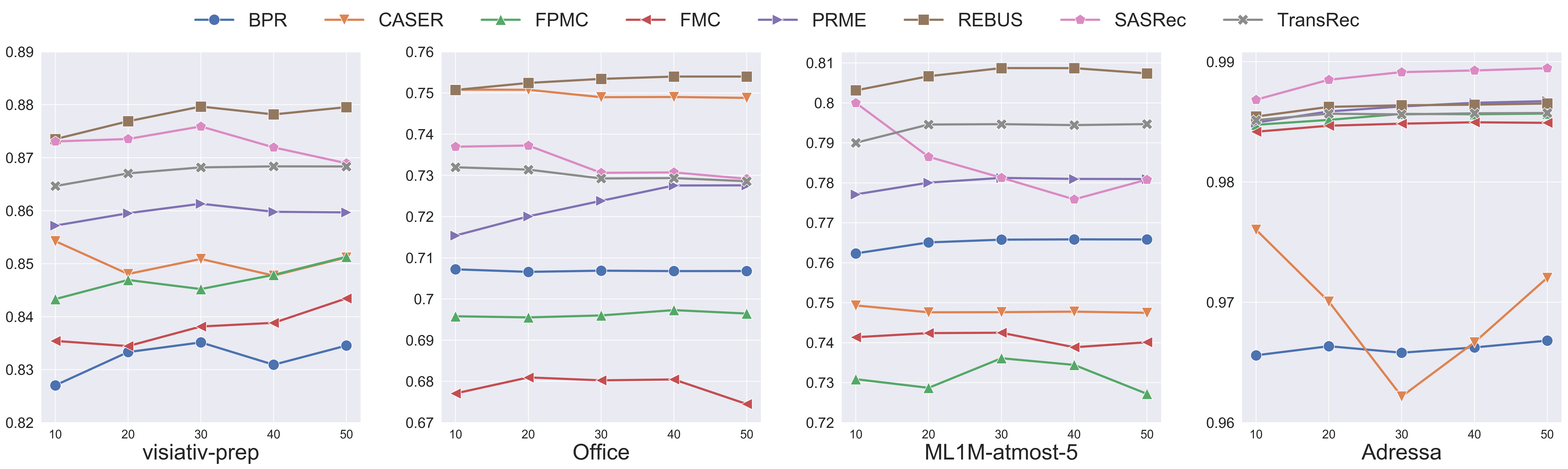}
  	\end{center}
	\caption{Influence on AUC (in ordinate) of the dimension $k$ (in abscissa) of the latent vectors ($k=\{10,20,30,40,50\}$).}
	\label{fig:xp_nums_dims}
\end{figure}

In Figure~\ref{fig:xp_nums_dims}, we study the effect of the size of the latent vectors ($k\in\{10,20,30,40,50\}$)\footnote{ To avoid  running again the grid search, we took the best combination of hyperparameters that we previously found for k = 10.} on AUC. In most of the cases, \algo{} has the best performances \modif{for each $k$ value.} It is worth to mention that \algo{} obtains better results when the size of the latent vectors increases, which is not the case of other methods. For instance, SASRec becomes worse when the number of latent dimensions increases for 3 of 4 datasets.

\begin{figure}[htb]
	\begin{center}
		\includegraphics[width=1\columnwidth]{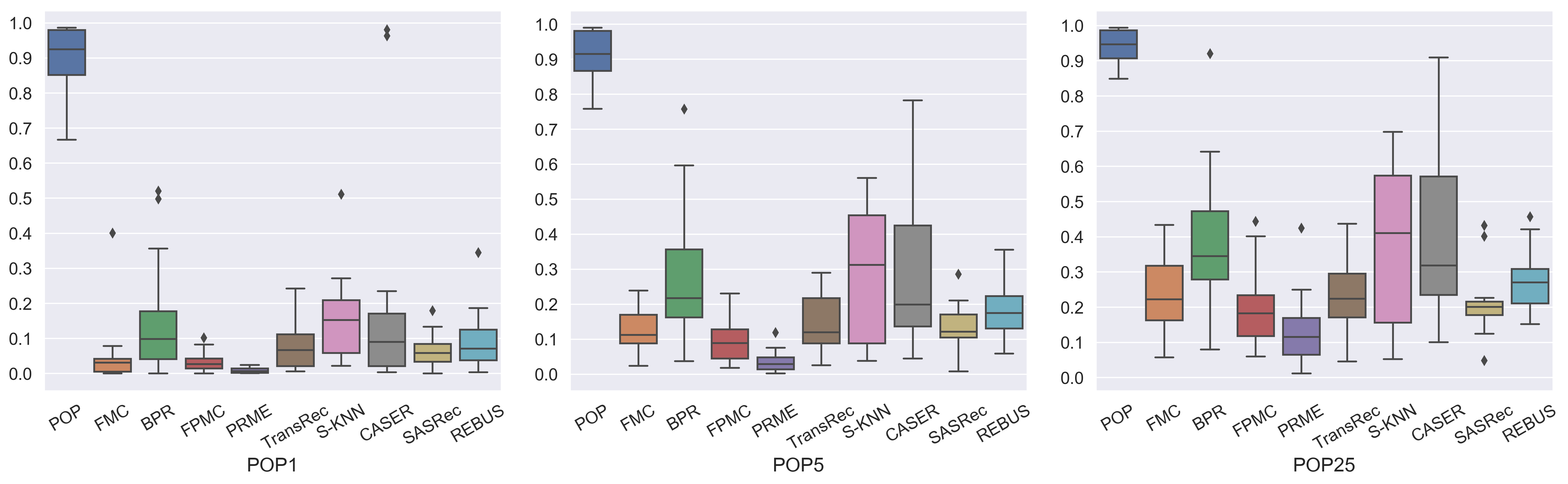}
		\includegraphics[width=1\columnwidth]{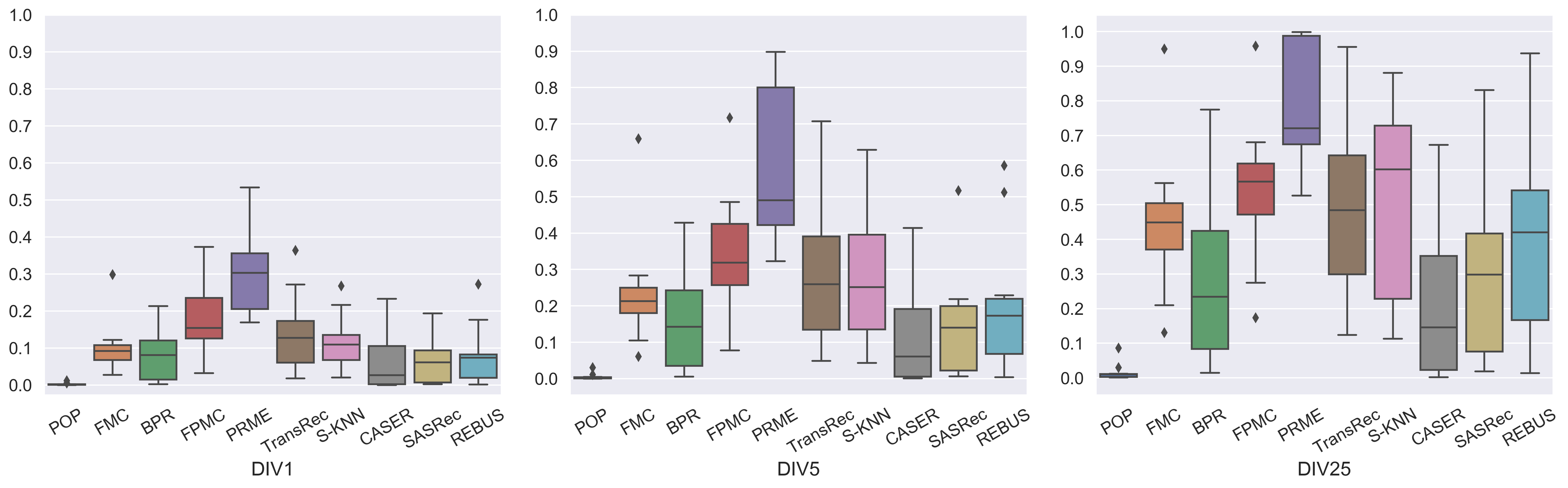}
	\end{center}
	\caption{\modif{Box plot of POP\_X (first row) and DIV\_X (second row) with $X\in\{1,5,25\}$ for each model. Values are averaged over all datasets.}}
	\label{fig:xp_preds_POP_COV}
\end{figure}

\modif{Eventually, we study the diversity of recommendations provided by each model. Figure~\ref{fig:xp_preds_POP_COV} reports POP\_X  and DIV\_X  for each model. 
For POP\_X, the higher the value, the more the model tends to recommend popular items. For DIV\_X the higher the value, the more the model tends to recommend different items.
These results give evidence that \algo{}' recommendations are rarely based on the most popular items. However, only a subset of items (between 20\% and 50\%) is recommended to the users, which remains comparable to most of the other models. }

\subsection{Cold-start user study}
We now investigate how \algo{} behaves when dealing with cold-start users, i.e., users whose sequence of actions is too short to take benefit from. We compare \algo{} to solely \modif{4 methods -- POP, FMC, SASRec and S-KNN -- } since the other ones are not applicable for cold-start user recommendation. Only models that just embed items can make recommendations in this context. To evaluate the performance of \algo{} and its competitors, we have chosen \modif{7} datasets: Epinion, Foursquare, \modif{Adressa},  Visiativ, Amazon-Automotive, Amazon-Office and Amazon-VideoGames. We then split each dataset into two parts:
\begin{enumerate}
\item One that contains users and items with at least 5 interactions. This is the same filter we used previously since we want to train the model with the same configuration and have the same hyperparameters;
\item A second one that contains cold-start users, that is to say users who do not appear in the first part but who interact with items that appear in the first part. Here, the users have a short history (between 1 and 4 items). This part is not used to train the models but only to evaluate their performance on the cold-start users.
\end{enumerate} 

For each dataset, the first parts are used as done in the previous studies. The second parts are split as follows: 
\begin{enumerate}
\item The most recent item is used for the test;
\item The other items of the user sequence are used to predict the test item.
\end{enumerate}

\begin{figure}[htb]
	\begin{center}
		\includegraphics[width=1\columnwidth]{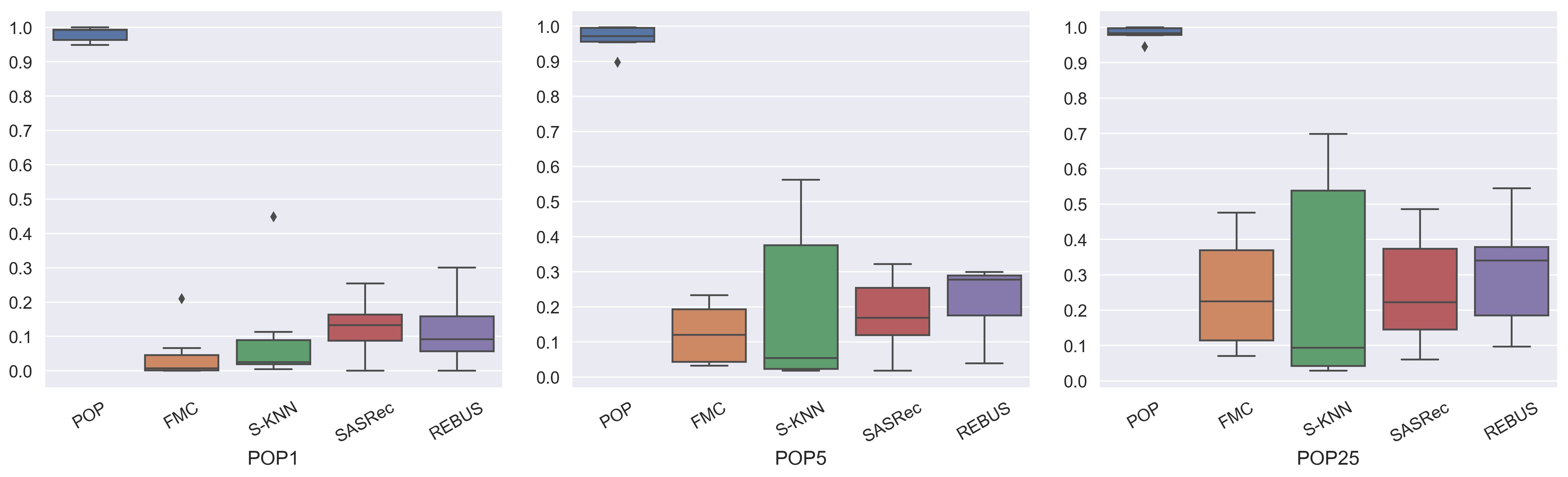}
		\includegraphics[width=1\columnwidth]{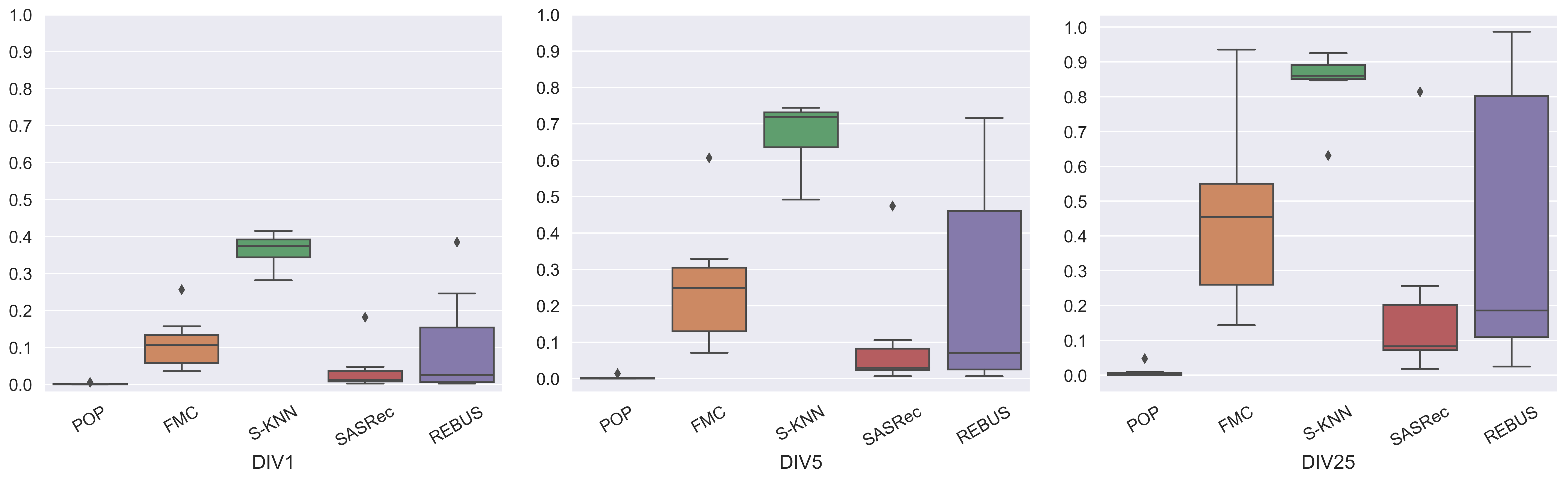}
	\end{center}
	\caption{\modif{Box plot of POP\_X (first row) and DIV\_X (second row) with $X\in\{1,5,25\}$ for the different models that do not suffer of the problem of cold-start users. Values are averaged over all datasets.}}
	\label{fig:xp_preds_POP_COV_cold_start}
\end{figure}

\modif{
The performances of the different methods on every dataset are reported in Table~\ref{tab:results_cold}\footnote{We only show HIT\_25, HIT\_50, NDCG\_25 and NDCG\_50 in the table. HIT\_5, HIT\_10, NDCG\_5 and NDCG\_10 are reported in supplementary material.}. We can observe that \algo{} outperforms other models except S-KNN on most of the datasets, having the best AUC and HIT\_50 values. S-KNN returns the best score for HIT\_25 and NDCG\_X. This can be explained by the fact that user sequences in this testbed are similar to those used in session-based recommendations (short sequences). In this context, S-KNN is well adapted for cold-start users. 
We report in Figure \ref{fig:xp_preds_POP_COV_cold_start} some additional indicators (i.e., POP\_X and DIV\_X) of the recommendations made for cold-start users. Results are similar to those obtained in Figure \ref{fig:xp_preds_POP_COV}. S-KNN provides more diverse recommendations than other models. This may explain it is the best model for HIT\_25 and NDCG\_X. 
This study demonstrates that \algo{} outperforms other models for the cold-start user problem for some performance metrics (i.e., AUC and HIT\_50). For others metrics, \algo{} is outperformed by only S-KNN. These good performances  are  mainly due to the embedding we use to model the data which is applicable on short user sequences.
}

\begin{table}[!h]
	\centering
	\caption{\modif{AUC, HIT\_25, HIT\_50, NDCG\_25 and NDCG\_50 for the different models that do not suffer of the problem of cold-start users. The 3 last rows, called \textit{Improv. vs Best}, \textit{Improv. vs SASRec} and \textit{Improv. vs S-KNN}, shows the improvement in percentage of our method compared to the best model, SASRec and S-KNN (best obtained results are in bold and best obtained results for concurrent models are underlined).}}
	\label{tab:results_cold}
	\scalebox{0.80}{
		\begin{tabular}{c|c|ccccccc|c}
			
			 & {\rotatebox{60}{\textbf{Metric}}} & {\rotatebox{60}{\textbf{Epinions}}} & {\rotatebox{60}{\textbf{Foursq}}} & {\rotatebox{60}{\textbf{Adressa}}} & {\rotatebox{60}{\textbf{Visiativ}}} & {\rotatebox{60}{\textbf{Auto}}} & {\rotatebox{60}{\textbf{Office}}} & {\rotatebox{60}{\textbf{Games}}} & {\rotatebox{60}{\textbf{Avg}}} \\\hline
			
			POP & AUC & 0.5741 & 0.8862 & 0.9689 & 0.7810 & 0.6400 & 0.6933 & 0.7702 & 0.7591 \\
			& HIT25 & \underline{3.62\%} & 41.38\% & 43.41\% & 31.21\% & 2.78\% & 0.46\% & 3.98\% & 18.12\% \\
			& HIT50 & \underline{5.41\%} & 51.93\% & 59.32\% & 46.08\% & 4.06\% & 1.69\% & 5.89\% & 24.91\% \\
			& NDGC25 & 1.41\% & 15.82\% & 18.55\% & 12.98\% & 1.09\% & 0.12\% & 1.55\% & 7.36\% \\
			& NDGC50 & 1.76\% & 17.88\% & 21.59\% & 15.81\% & 1.34\% & 0.36\% & 1.91\% & 8.66\% \\
			FMC & AUC & 0.5829 & 0.9326 & 0.9776 & 0.8316 & 0.6644 & 0.7020 & 0.8529 & 0.7920 \\
			& HIT25 & 2.28\% & 48.89\% & 59.37\% & 45.54\% & 3.32\% & 1.32\% & 10.88\% & 24.51\% \\
			& HIT50 & 3.61\% & 60.15\% & 72.33\% & 57.97\% & 4.84\% & 2.72\% & 15.63\% & 31.04\% \\
			& NDGC25 & 0.91\% & 22.98\% & 27.35\% & 23.84\% & 1.50\% & 0.54\% & 4.27\% & 11.63\% \\
			& NDGC50 & 1.16\% & 25.16\% & 29.84\% & 26.22\% & 1.79\% & 0.81\% & 5.19\% & 12.88\% \\
			S-KNN & AUC & 0.0495 & 0.7012 & 0.9515 & 0.8048 & 0.1138 & 0.1297 & 0.4103 & 0.4515 \\
			& HIT25 & 3.09\% & \underline{\textbf{54.48\%}} & 58.82\% & \underline{51.01\%} & \underline{\textbf{6.91\%}} & \underline{\textbf{7.20\%}} & \underline{\textbf{14.83\%}} & \underline{\textbf{28.05\%}} \\
			& HIT50 & 3.90\% & \underline{61.65\%} & 70.61\% & 61.18\% & \underline{\textbf{8.21\%}} & \underline{\textbf{8.62\%}} & \underline{\textbf{18.93\%}} & \underline{33.30\%} \\
			& NDGC25 & \underline{\textbf{1.74\%}} & \underline{\textbf{34.67\%}} & \underline{\textbf{34.60\%}} & \underline{\textbf{27.89\%}} & \underline{\textbf{3.68\%}} & \underline{\textbf{4.15\%}} & \underline{\textbf{6.91\%}} & \underline{\textbf{16.24\%}} \\
			& NDGC50 & \underline{1.90\%} & \underline{\textbf{36.06\%}} & \underline{\textbf{36.87\%}} & \underline{\textbf{29.84\%}} & \underline{\textbf{3.93\%}} & \underline{\textbf{4.43\%}} & \underline{\textbf{7.70\%}} & \underline{\textbf{17.25\%}} \\
			SASRec & AUC & \underline{\textbf{0.6504}} & \underline{0.9407} & \underline{0.9806} & \underline{\textbf{0.8655}} & \underline{0.6893} & \underline{\textbf{0.7340}} & \underline{0.8698} & \underline{0.8186} \\
			& HIT25 & 3.36\% & 47.57\% & \underline{61.29\%} & 49.11\% & 2.85\% & 2.96\% & 9.26\% & 25.20\% \\
			& HIT50 & 5.07\% & 58.44\% & \underline{74.98\%} & \underline{61.59\%} & 4.20\% & 4.84\% & 14.12\% & 31.89\% \\
			& NDGC25 & 1.29\% & 23.12\% & 29.14\% & 24.56\% & 1.12\% & 1.05\% & 3.66\% & 11.99\% \\
			& NDGC50 & 1.61\% & 25.22\% & 31.77\% & 26.97\% & 1.38\% & 1.41\% & 4.60\% & 13.28\% \\ \hline
			
			REBUS & AUC & 0.6253 & \textbf{0.9537} & \textbf{0.9815} & 0.8612 & \textbf{0.7267} & 0.7309 & \textbf{0.8747} & \textbf{0.8220} \\
			& HIT25 & \textbf{3.94\%} & 53.75\% & \textbf{63.42\%} & \textbf{51.61\%} & 4.57\% & 5.09\% & 10.46\% & 27.55\% \\
			& HIT50 & \textbf{6.01\%} & \textbf{64.47\%} & \textbf{75.65\%} & \textbf{62.78\%} & 6.76\% & 7.65\% & 15.88\% & \textbf{34.17\%} \\
			& NDGC25 & 1.58\% & 30.21\% & 33.22\% & 26.23\% & 1.79\% & 2.07\% & 4.04\% & 14.16\% \\
			& NDGC50 & \textbf{1.97\%} & 32.28\% & 35.58\% & 28.39\% & 2.21\% & 2.56\% & 5.08\% & 15.44\% \\ \hline
			
			Rank & AUC & 2 & 1 & 1 & 2 & 1 & 2 & 1 & 1.43 \\
			& HIT25 & 1 & 2 & 1 & 1 & 2 & 2 & 3 & 1.71 \\
			& HIT50 & 1 & 1 & 1  & 1 & 2 & 2 & 2 & 1.43 \\
			& NDGC25 & 2 & 2 & 2 & 2 & 2 & 2 & 3 & 2.14 \\
			& NDGC50 & 1 & 2 & 2 & 2 & 2 & 2 & 3 & 2.00 \\ \hline

			Improv. & AUC & -3.86\% & 1.38\% & 0.09\% & -0.50\% & 5.43\% & -0.42\% & 0.56\% & 0.42\% \\
			VS & HIT25 & 8.84\% & -1.34\% & 3.48\% & 1.18\% & -33.86\% & -29.31\% & -29.47\% & -1.78\% \\
			Best & HIT50 & 11.09\% & 4.57\% & 0.89\% & 1.93\% & -17.66\% & -11.25\% & -16.11\% & 2.61\% \\
			& NDGC25 & -9.20\% & -12.86\% & -3.99\% & -5.95\% & -51.36\% & -50.12\% & -41.53\% & -12.81\% \\
			& NDGC50 & 3.68\% & -10.48\% & -3.50\% & -4.86\% & -43.77\% & -42.21\% & -34.03\% & -10.49\% \\
			Improv. & AUC & -3.86\% & 1.38\% & 0.09\% & -0.50\% & 5.43\% & -0.42\% & 0.56\% & 0.42\% \\
			VS & HIT25 & 17.26\% & 12.99\% & 3.48\% & 5.09\% & 60.35\% & 71.96\% & 12.96\% & 9.33\% \\
			SASRec & HIT50 & 18.54\% & 10.32\% & 0.89\% & 1.93\% & 60.95\% & 58.06\% & 12.46\% & 7.15\% \\
			& NDGC25 & 22.48\% & 30.67\% & 14.00\% & 6.80\% & 59.82\% & 97.14\% & 10.38\% & 18.10\% \\
			& NDGC50 & 22.36\% & 27.99\% & 11.99\% & 5.27\% & 60.14\% & 81.56\% & 10.43\% & 16.27\% \\
			Improv. & AUC & 1163\% & 36.01\% & 3.15\% & 7.01\% & 538.58\% & 463.53\% & 113.19\% & 82.06\% \\
			VS & HIT25 & 27.51\% & -1.34\% & 7.82\% & 1.18\% & -33.86\% & -29.31\% & -29.47\% & -1.78\% \\
			 S-KNN & HIT50 & 54.10\% & 4.57\% & 7.14\% & 2.62\% & -17.66\% & -11.25\% & -16.11\% & 2.61\% \\
			& NDGC25 & -9.20\% & -12.86\% & -3.99\% & -5.95\% & -51.36\% & -50.12\% & -41.53\% & -12.81\% \\
			& NDGC50 & 3.68\% & -10.48\% & -3.50\% & -4.86\% & -43.77\% & -42.21\% & -34.03\% & -10.49\% \\ 
			\hline
			
		\end{tabular}
	}
\end{table}

\subsection{Study of the impact of user preferences and sequential dynamics in the recommendation}

\begin{table}[htb]
	\centering
	\caption{AUC, HIT\_25, HIT\_50, NDCG\_25 and NDCG\_50 for \algo{} on \modif{Epinions, Foursquare, Adressa, Visiativ, Amazon-Automotive, Amazon-Office-Product and Amazon-Video-Games datasets} (best obtained results are in bold). The last column is the average performance on these 7 datasets, the average of all datasets is included in Table~\ref{tab:results_ablationBIS}.}
	\label{tab:results_ablation}
	\scalebox{0.8}{
		\rotatebox{0}{
			\begin{tabular}{c|c|ccccccc|c}
			  & {\rotatebox{60}{\textbf{Metric}}} & {\rotatebox{60}{\textbf{Epinions}}} & {\rotatebox{60}{\textbf{Foursq}}} & {\rotatebox{60}{\textbf{Adressa}}} & {\rotatebox{60}{\textbf{Visiativ}}} & {\rotatebox{60}{\textbf{Auto}}} & {\rotatebox{60}{\textbf{Office}}} & {\rotatebox{60}{\textbf{Games}}} & {\rotatebox{60}{\textbf{Avg}}} \\\hline
			 
				
				\multirow{5}{*}{\rotatebox{60}{\tiny{\algoLT}}}
				& AUC & 0.6495 & 0.9661 & 0.9828 & 0.8521 & 0.7147 & 0.7484 & 0.8860 & 0.8285 \\
				& HIT25 & 3.21\% & 61.16\% & 63.48\% & 43.80\% & 4.08\% & 3.88\% & 8.83\% & 26.92\% \\
				& HIT50 & 5.00\% & 69.20\% & 76.93\% & 59.64\% & 6.02\% & 6.06\% & 14.14\% & 33.86\% \\
				& NDGC25 & 1.27\% & 42.07\% & 31.80\% & 18.29\% & 1.62\% & 1.53\% & 3.28\% & 14.26\% \\
				& NDGC50 & 1.61\% & 43.62\% & 34.40\% & 21.34\% & 1.99\% & 1.94\% & 4.30\% & 15.60\% \\ \hline
				
				\multirow{5}{*}{\rotatebox{60}{\tiny{\algoST$_{1MC}$}}}
				& AUC & 0.6257 & 0.9624 & 0.9846 & 0.8591 & 0.6855 & 0.7184 & 0.8710 & 0.8153\\
				& HIT25 & 2.65\% & 55.12\% & 64.26\% & 50.97\% & 3.80\% & 4.08\% & 9.16\% & 27.15\% \\
				& HIT50 & 3.84\% & 65.09\% & 77.89\% & 63.58\% & 5.55\% & 6.74\% & 13.74\% & 33.78\% \\
				& NDGC25 & 1.19\% & 28.93\% & 30.23\% & \textbf{27.11\%} & 1.47\% & 1.52\% & \textbf{3.59\%} & 13.44\%\\
				& NDGC50 & 1.42\% & 30.85\% & 32.87\% & \textbf{29.55\%} & 1.81\% & 2.03\% & 4.47\% & 14.71\% \\
				\multirow{5}{*}{\rotatebox{60}{\tiny{\algoST$_{2MC}$}}}
				& AUC & 0.6428 & 0.9660 & 0.9852 & 0.8720 & 0.7132 & 0.7440 & 0.8872 & 0.8301 \\
				& HIT25 & 3.00\% & 55.26\% & 64.75\% & \textbf{53.55\%} & 4.11\% & 4.40\% & 9.01\% & 27.73\% \\
				& HIT50 & 5.30\% & 65.07\% & 78.66\% & 66.45\% & 5.96\% & 6.93\% & 13.95\% & 34.62\% \\
				& NDGC25 & 1.29\% & 30.11\% & 30.67\% & 25.82\% & 1.61\% & 1.59\% & 3.47\% & 13.51\% \\
				& NDGC50 & 1.73\% & 32.01\% & 33.36\% & 28.31\% & 1.96\% & 2.08\% & 4.42\% & 14.84\% \\
				\multirow{5}{*}{\rotatebox{60}{\tiny{\algoST$_{3MC}$}}}
				& AUC & 0.6485 & 0.9663 & 0.9850 & 0.8725 & \textbf{0.7192} & 0.7536 & 0.8892 & 0.8335 \\
				& HIT25 & 3.35\% & 55.21\% & 64.56\% & 52.90\% & 4.15\% & \textbf{4.60\%} & 8.68\% & 27.64\% \\
				& HIT50 & 5.30\% & 64.92\% & 78.17\% & 66.31\% & 6.15\% & \textbf{7.31\%} & 13.57\% & 34.53\% \\
				& NDGC25 & 1.34\% & 30.91\% & 30.36\% & 24.23\% & 1.60\% & \textbf{1.66\%} & 3.36\% & 13.35\% \\
				& NDGC50 & 1.71\% & 32.79\% & 32.99\% & 26.81\% & 1.99\% & \textbf{2.18\%} & 4.30\% & 14.68\% \\ \hline
				
				\multirow{5}{*}{\rotatebox{60}{\tiny{\algoST}}}
				& AUC & 0.6114 & 0.9653 & 0.9849 & 0.8610 & 0.6904 & 0.7232 & 0.8796 & 0.8166 \\
				& HIT25 & 3.02\% & 55.37\% & 64.78\% & 51.25\% & 3.89\% & 4.11\% & 9.09\% & 27.36\% \\
				& HIT50 & 4.25\% & 65.20\% & 78.41\% & 63.80\% & 5.84\% & 6.69\% & 13.78\% & 34.00\% \\
				& NDGC25 & 1.20\% & 30.72\% & 30.47\% & 26.35\% & 1.51\% & 1.49\% & 3.53\% & 13.61\% \\
				& NDGC50 & 1.43\% & 32.62\% & 33.10\% & 28.78\% & 1.89\% & 1.99\% & 4.43\% & 14.89\% \\ \hline
				
				\multirow{5}{*}{\rotatebox{60}{\tiny{\algo$_{1MC}$}}}
				& AUC & 0.6584 & \textbf{0.9682} & 0.9854 & 0.8751 & 0.7190 & \textbf{0.7542} & 0.8935 & 0.8363 \\
				& HIT25 & \textbf{3.67\%} & \textbf{63.21\%} & 66.45\% & 53.48\% & 4.11\% & 4.27\% & 9.28\% & \textbf{29.21\%} \\
				& HIT50 & 5.18\% & \textbf{70.81\%} & 79.71\% & \textbf{67.60\%} & 5.99\% & 6.44\% & \textbf{14.73\%} & \textbf{35.78\%} \\
				& NDGC25 & \textbf{1.46\%} & \textbf{46.16\%} & 31.77\% & 24.59\% & 1.62\% & 1.55\% & 3.53\% & \textbf{15.81\%} \\
				& NDGC50 & \textbf{1.75\%} & \textbf{47.63\%} & 34.33\% & 27.31\% & 1.99\% & 1.97\% & \textbf{4.57\%} & \textbf{17.08\%} \\
				\multirow{5}{*}{\rotatebox{60}{\tiny{\algo$_{2MC}$}}}
				& AUC & \textbf{0.6660} & 0.9676 & \textbf{0.9855} & \textbf{0.8754} & 0.7189 & 0.7537 & \textbf{0.8936} & \textbf{0.8372} \\
				& HIT25 & 3.25\% & 62.76\% & 66.61\% & 52.40\% & 4.16\% & 4.15\% & 9.17\% & 28.93\% \\
				& HIT50 & \textbf{5.56\%} & 70.26\% & 79.68\% & 66.16\% & 6.07\% & 6.51\% & 14.55\% & 35.54\% \\
				& NDGC25 & 1.31\% & 45.92\% & 31.92\% & 23.58\% & 1.62\% & 1.60\% & 3.46\% & 15.63\% \\
				& NDGC50 & \textbf{1.75\%} & 47.37\% & 34.44\% & 26.21\% & 1.99\% & 2.05\% & 4.49\% & 16.90\% \\
				\multirow{5}{*}{\rotatebox{60}{\tiny{\algo$_{3MC}$}}}
				& AUC & 0.6624 & 0.9669 & 0.9849 & 0.8729 & 0.7179 & 0.7528 & 0.8923 & 0.8357 \\
				& HIT25 & 3.49\% & 62.92\% & 64.60\% & 50.39\% & 4.13\% & 4.19\% & 8.82\% & 28.36\% \\
				& HIT50 & 5.44\% & 70.24\% & 78.53\% & 64.95\% & 6.01\% & 6.32\% & 14.06\% & 35.08\% \\
				& NDGC25 & 1.35\% & 46.14\% & 30.36\% & 22.57\% & 1.62\% & 1.61\% & 3.31\% & 15.28\% \\
				& NDGC50 & 1.73\% & 47.55\% & 33.05\% & 25.36\% & 1.98\% & 2.02\% & 4.32\% & 16.57\% \\ \hline
				
				\multirow{5}{*}{\rotatebox{60}{\tiny{\algo}}}
				& AUC & 0.6524 & 0.9677 & 0.9854 & 0.8735 & 0.7184 & 0.7507 & 0.8934 & 0.8345 \\
				& HIT25 & 3.39\% & 62.77\% & \textbf{66.73\%} & 53.33\% & \textbf{4.24\%} & 4.12\% & \textbf{9.30\%} & 29.13\% \\
				& HIT50 & 5.32\% & 70.40\% & \textbf{79.77\%} & 67.17\% & \textbf{6.20\%} & 6.24\% & 14.61\% & 35.67\% \\
				& NDGC25 & 1.31\% & 46.04\% & \textbf{32.00\%} & 24.20\% & \textbf{1.68\%} & 1.52\% & 3.52\% & 15.75\% \\
				& NDGC50 & 1.68\% & 47.52\% & \textbf{34.52\%} & 26.86\% & \textbf{2.05\%} & 1.92\% & 4.54\% & 17.01\% \\ \hline
				
			\end{tabular}
		}
	}
\end{table}

\begin{table}[htb]
	\centering
	\caption{\modif{AUC, HIT\_25, HIT\_50, NDCG\_25 and NDCG\_50 for \algo{} on ML-5, ML-10, ML-20, ML-30 and ML-50 datasets (best obtained results are in bold). The last column is the average performance on the 12 datasets, including datasets of Table~\ref{tab:results_ablation}.}}
	\label{tab:results_ablationBIS}
	\scalebox{0.8}{
		\rotatebox{0}{
			\begin{tabular}{c|c|cccccc|c}
				
				 & {\rotatebox{60}{\textbf{Metric}}} & {\rotatebox{60}{\textbf{ML-5}}} & {\rotatebox{60}{\textbf{ML-10}}} & {\rotatebox{60}{\textbf{ML-20}}} & {\rotatebox{60}{\textbf{ML-30}}} & {\rotatebox{60}{\textbf{ML-50}}} & {\rotatebox{60}{\textbf{Avg(ML)}}} & {\rotatebox{60}{\textbf{Avg(All)}}} \\ \hline
				
				\multirow{5}{*}{\rotatebox{60}{\tiny{\algoLT}}}
				& AUC & 0.7941 & 0.8428 & 0.8603 & 0.8631 & 0.8671 & 0.8455 & 0.8356 \\
				& HIT25 & 12.95\% & 12.49\% & 11.97\% & 11.61\% & 11.39\% & 12.08\% & 20.74\% \\
				& HIT50 & 21.51\% & 20.72\% & 19.99\% & 19.47\% & 19.49\% & 20.24\% & 28.18\% \\
				& NDGC25 & 4.80\% & 4.40\% & 4.10\% & 4.06\% & 3.88\% & 4.25\% & 10.09\% \\
				& NDGC50 & 6.43\% & 5.97\% & 5.64\% & 5.56\% & 5.43\% & 5.81\% & 11.52\% \\ \hline
								
				\multirow{5}{*}{\rotatebox{60}{\tiny{\algoST$_{1MC}$}}}
				& AUC & 0.7848 & 0.8207 & 0.8556 & 0.8657 & 0.8725 & 0.8399 & 0.8255 \\ 
				& HIT25 & 13.45\% & 15.66\% & 17.83\% & 18.08\% & 18.20\% & 16.65\% & 22.77\% \\
				& HIT50 & 20.83\% & 23.22\% & 26.38\% & 27.14\% & 27.39\% & 24.99\% & 30.12\% \\
				& NDGC25 & 5.10\% & 6.18\% & 6.74\% & 6.99\% & 7.03\% & 6.41\% & 10.51\% \\
				& NDGC50 & 6.52\% & 7.63\% & 8.39\% & 8.73\% & 8.80\% & 8.01\% & 11.92\% \\
				\multirow{5}{*}{\rotatebox{60}{\tiny{\algoST$_{2MC}$}}}
				& AUC &  0.7954 & 0.8468 & 0.8718 & 0.8773 & 0.8850 & 0.8553 & 0.8406 \\
				& HIT25 & 14.06\% & \textbf{18.15\%} & 19.32\% & \textbf{19.71\%} & \textbf{19.75\%} & \textbf{18.20\%} & \textbf{23.76\%} \\
				& HIT50 & 22.07\% & 26.89\% & 29.03\% & \textbf{30.02\%} & 30.29\% & 27.66\% & 31.72\% \\
				& NDGC25 & 5.17\% & \textbf{6.75\%} & 7.28\% & \textbf{7.74\%} & \textbf{7.49\%} & \textbf{6.88\%} & 10.75\% \\
				& NDGC50 & 6.71\% & 8.43\% & 9.14\% & \textbf{9.71\%} & \textbf{9.50\%} & \textbf{8.70\%} & 12.28\% \\
				\multirow{5}{*}{\rotatebox{60}{\tiny{\algoST$_{3MC}$}}}
				& AUC &  0.7969 & 0.8531 & 0.8763 & 0.8804 & 0.8879 & 0.8589 & 0.8441 \\
				& HIT25 & 13.78\% & 17.34\% & \textbf{19.56\%} & 19.44\% & 19.47\% & 17.92\% & 23.59\% \\
				& HIT50 & 21.87\% & \textbf{27.57\%} & \textbf{29.92\%} & 29.86\% & \textbf{30.44\%} & \textbf{27.93\%} & \textbf{31.78\%} \\
				& NDGC25 & 5.03\% & 6.48\% & \textbf{7.35\%} & 7.51\% & 7.28\% & 6.73\% & 10.59\% \\
				& NDGC50 & 6.58\% & \textbf{8.44\%} & \textbf{9.34\%} & 9.51\% & 9.39\% & 8.65\% & 12.17\% \\ \hline
					
				\multirow{5}{*}{\rotatebox{60}{\tiny{\algoST}}}
				& AUC & 0.7758 & 0.8282 & 0.8572 & 0.8677 & 0.8756 & 0.8409 & 0.8267 \\ 
				& HIT25 & 13.96\% & 16.05\% & 17.50\% & 18.26\% & 18.43\% & 16.84\% & 22.98\% \\
				& HIT50 & 21.29\% & 23.88\% & 26.73\% & 27.77\% & 27.79\% & 25.49\% & 30.45\% \\
				& NDGC25 & 5.16\% & 6.13\% & 6.49\% & 7.03\% & 6.83\% & 6.33\% & 10.58\% \\
				& NDGC50 & 6.57\% & 7.64\% & 8.26\% & 8.86\% & 8.62\% & 7.99\% & 12.02\% \\ \hline
					
				\multirow{5}{*}{\rotatebox{60}{\tiny{\algo$_{1MC}$}}}
				& AUC & 0.8023 & 0.8555 & 0.8770 & 0.8829 & 0.8883 & 0.8612 & 0.8466 \\
				& HIT25 & \textbf{14.82\%} & 15.17\% & 15.80\% & 16.58\% & 16.87\% & 15.85\% & 23.64\% \\
				& HIT50 & 22.67\% & 24.66\% & 26.61\% & 27.21\% & 27.34\% & 25.70\% & 31.58\% \\
				& NDGC25 & \textbf{5.44\%} & 5.48\% & 5.68\% & 6.08\% & 6.21\% & 5.78\% & \textbf{11.63\%} \\
				& NDGC50 & \textbf{6.95\%} & 7.30\% & 7.76\% & 8.12\% & 8.21\% & 7.67\% & \textbf{13.16\%} \\
				\multirow{5}{*}{\rotatebox{60}{\tiny{\algo$_{2MC}$}}}
				& AUC & 0.8012 & 0.8557 & 0.8770 & \textbf{0.8837} & \textbf{0.8905} & \textbf{0.8616} & \textbf{0.8474} \\
				& HIT25 & 14.46\% & 14.85\% & 15.35\% & 16.48\% & 15.91\% & 15.41\% & 23.30\% \\
				& HIT50 & 22.31\% & 24.82\% & 25.53\% & 27.17\% & 26.78\% & 25.32\% & 31.28\% \\
				& NDGC25 & 5.24\% & 5.38\% & 5.45\% & 6.04\% & 5.83\% & 5.59\% & 11.45\% \\
				& NDGC50 &  6.74\% & 7.29\% & 7.40\% & 8.09\% & 7.92\% & 7.49\% & 12.98\% \\
				\multirow{5}{*}{\rotatebox{60}{\tiny{\algo$_{3MC}$}}}
				& AUC &  0.8000 & 0.8538 & 0.8764 & 0.8825 & 0.8894 & 0.8604 & 0.8460 \\
				& HIT25 & 14.03\% & 14.41\% & 14.82\% & 15.78\% & 15.63\% & 14.93\% & 22.77\% \\
				& HIT50 &  21.96\% & 24.06\% & 25.30\% & 26.31\% & 25.62\% & 24.65\% & 30.73\% \\
				& NDGC25 & 5.08\% & 5.25\% & 5.29\% & 5.67\% & 5.53\% & 5.36\% & 11.15\% \\
				& NDGC50 & 6.61\% & 7.10\% & 7.29\% & 7.69\% & 7.45\% & 7.23\% & 12.68\% \\ \hline
					
				\multirow{5}{*}{\rotatebox{60}{\tiny{\algo}}}
				& AUC & \textbf{0.8031} & \textbf{0.8563} & \textbf{0.8772} & 0.8826 & 0.8884 & 0.8615 & 0.8458 \\ 
				& HIT25 & 14.42\% & 15.27\% & 15.60\% & 15.80\% & 16.72\% & 15.56\% & 23.48\% \\
				& HIT50 & \textbf{22.79\%} & 25.50\% & 26.01\% & 26.21\% & 26.91\% & 25.48\% & 31.43\% \\
				& NDGC25 & 5.32\% & 5.59\% & 5.60\% & 5.80\% & 6.03\% & 5.66\% & 11.55\% \\
				& NDGC50 & 6.92\% & 7.55\% & 7.59\% & 7.80\% & 7.98\% & 7.57\% & 13.08\% \\ \hline
			\end{tabular}
		}
	}
\end{table}

We have demonstrated that \algo{} outperforms the state-of-the-art algorithms in most of the configurations (i.e., datasets and metrics). We now study how our model effectively behave. Especially, we investigate the impact of both user preferences and sequential dynamics on recommendation. To this end, we consider independently these two components. Furthermore, we also examine short Markov chains in our model instead of frequent sequences to capture the sequential dynamics. \algoLT{} refers to the long term part, \algoST{} refers to the short term part using personalized context via frequent sequences, ${XMC}${} means that the model uses Markov chains with a fixed order $X$. The performances of the different configurations on each dataset are reported in Tables~\ref{tab:results_ablation} and ~\ref{tab:results_ablationBIS}.

As expected, we can observe that the configurations of our model that combine user preferences and sequential dynamics (i.e. \algo{} and \algo$_{XMC}$) outperform other configurations -- that only model either user preferences or the sequential dynamics -- (1) for all metrics on sparse datasets and (2) only for AUC on dense datasets. Indeed, surprisingly, \algoST$_{XMC}$ (when $X>1$) outperforms on dense datasets all models tested in our experimental study, SASRec included, for HIT\_X and NDGC\_X (see Table~\ref{tab:resultsbis}). Despite the fact that \algoST$_{XMC}$ (when $X>1$) outperforms all non-combining configurations (that is to say \algoST$_{1MC}$, \algoST{} and \algoLT),
there is no improvement when $XMC$ is used in \algo{} (i.e. \algo$_{XMC}$ with $X>1$).
In overall, \algoST{} provides better performances than \algoST$_{1MC}$, which proves that there is an interest in using personalized sequences to model sequential dynamics. To conclude, the most consistent configurations are \algo$_{1MC}$ and \algo.

\subsection{Study of the used personalized sequences} \label{subsec:xp_contexts}

We study the characteristics of the sequences used by \algo{} for recommendation. The first five columns (from (A) to (E)) of Table~\ref{tab:nb_root} \footnote{The two numbers after the name of the dataset are respectively the value of \minCount{} and \maxSize.} report the percentage of sequences $\frequ$ that are equivalent to: (A) no matched items, (B) a first order Markov chain \footnote{Column (A) and (B) are both equivalent to a first order Markov chain but we split it into two different columns to point out the fact that it is very uncommon that no item from \freq{} matches $s_{u}^{[1,t[}$}, (C) a first order Markov chain but not with the most recent item (D) a $L$ order Markov chain ($L>1$). In column (E) we report the percentage of sequences $\frequ$ that cannot be modeled by Markov chains (sequences that are not strings). The column (A) confirms that it is very uncommon that none of the sequences of \freq{} match the sequence $s_{u}^{[1,t[}$. \algo{} often takes into account the most recent item within the user's sequence, but there are some cases for which older items are used. This cannot be obtained by first order Markov chains. Furthermore, depending on the dataset, between 10\% to 40\% (Columns (C) + (E)) of the sequences are different from Markov Chains. The added value of our approach can be assessed by (C), (D) and (E). The two last columns\footnote{Sequences that do not match any substring of \freq{} are omitted for columns (F) and (G).} of Table~\ref{tab:nb_root} reports the mean of $\frequ$ sizes (F) and the mean of their occupation in $s_u$ (G), i.e., difference between the last position and the first position in $\frequ$. 
In most cases, short sequences are exploited to model the sequential dynamics. Nevertheless, we have observed that there are different sizes ranging from 1 to 15. The mean of the occupation of the sequences is rather low. \algo{} often uses \emph{compact} sequences. This can explain the good results of short Markov chains to capture sequential dynamics in \algo. We can also observe that \algo{} sometimes uses non-consecutive items with an arbitrary number of spaces between items. Such sequences cannot be obtained by fixed order Markov chains. This is why frequent sequences outperforms Markov chain when focusing only on the sequential dynamics (see Table~\ref{tab:results_ablation}).

\begin{table}[htb]
	\begin{center}
		\caption{The first five columns show the percentage of substrings $\frequ$ that are  equivalent to: (A) no matched items, (B) a first order Markov chain, (C) a first order Markov chain but not with the most recent item, (D) a $L$ order Markov chain ($L>1$). (E) shows the percentage of substrings $\frequ$ that cannot be modeled by Markov chains (substrings that are mapped in a non-consecutive way on the user history). The two last columns show the mean of $\frequ$ sizes (F) and the mean of their occupation length in the user sequence (G).}
		\label{tab:nb_root}
		\scalebox{0.95}{
			\begin{tabular}{|c|c|c|c|c|c|c|c|}
				\hline
				Dataset           & \textbf{(A)} & \textbf{(B)} & \textbf{(C)} & \textbf{(D)} & \textbf{(E)} & \textbf{(F)} & \textbf{(G)}    \\
				& no match     & MC$_1$       &  MC$_1$\_old &  MC$_L$ & Seq.         & Size         & Occup.     \\ \hline
				Epinions\_2\_3    & 3.39\%       & 62.21\%      & 34.06\%      & 0.16\%  & 0.18\%       & 1.01      & 1.01   \\
				Foursquare\_2\_15 & 0.06\%       & 26.15\%      & 1.39\%       & 43.85\% & 28.54\%      & 2.36      & 2.93   \\
				Adressa\_2\_2     & 0.00\%       & 1.66\%       & 0.02\%       & 93.05\% & 5.26\%       & 1.98      & 2.07   \\
				Visiativ\_2\_10   & 0.00\%       & 40.27\%      & 0.21\%       & 28.97\% & 30.54\%      & 1.71      & 2.91   \\
				Ama-Auto\_2\_8    & 2.12\%       & 72.82\%      & 22.48\%      & 1.31\%  & 1.27\%       & 1.03      & 1.04   \\
				Ama-Office\_2\_8  & 1.19\%       & 72.86\%      & 20.79\%      & 2.69\%  & 2.48\%       & 1.05      & 1.10   \\
				Ama-Games\_2\_10  & 0.06\%       & 80.74\%      & 5.57\%       & 6.13\%  & 7.50\%       & 1.14      & 1.38   \\
				ML-5\_2\_3        & 0.00\%       & 87.05\%      & 7.50\%       & 3.08 \% & 2.37\%       & 1.06      & 1.08   \\
				ML-10\_2\_8       & 0.00\%       & 79.35\%      & 1.94\%       & 8.06 \% & 10.65\%      & 1.20      & 1.48   \\
				ML-20\_2\_8       & 0.00\%       & 59.11\%      & 0.81\%       & 13.41\% & 26.67\%      & 1.42      & 2.90   \\
				ML-30\_2\_10      & 0.00\%       & 47.37\%      & 0.51\%       & 16.71\% & 35.41\%      & 1.55      & 4.21   \\
				ML-50\_2\_10      & 0.00\%       & 35.33\%      & 0.17\%       & 20.94\% & 43.56\%      & 1.68      & 5.96   \\ \hline
			\end{tabular}
		}
	\end{center}
\end{table}

\subsection{Relative importance of user preferences on sequential dynamics}
\begin{figure}[!htb]
	\includegraphics[width=1\columnwidth]{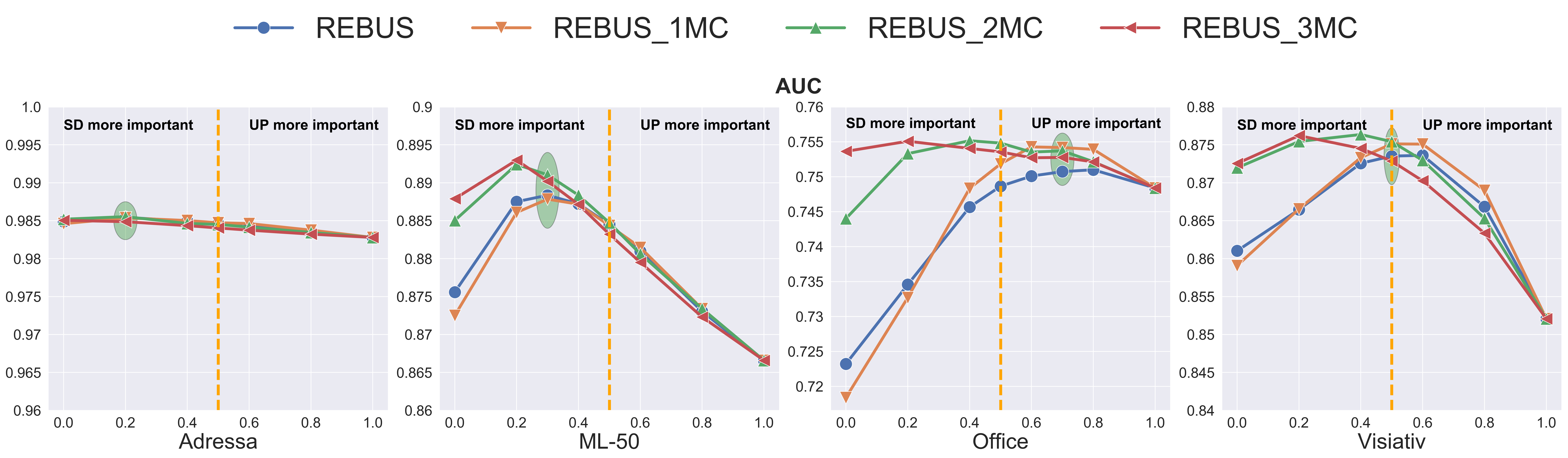}
	\includegraphics[width=1\columnwidth]{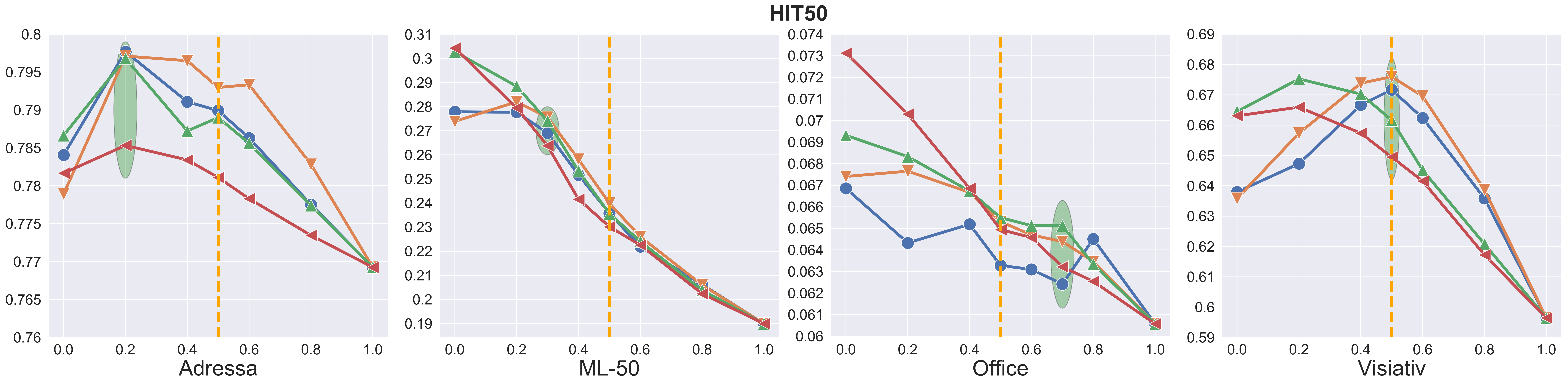}
	\includegraphics[width=1\columnwidth]{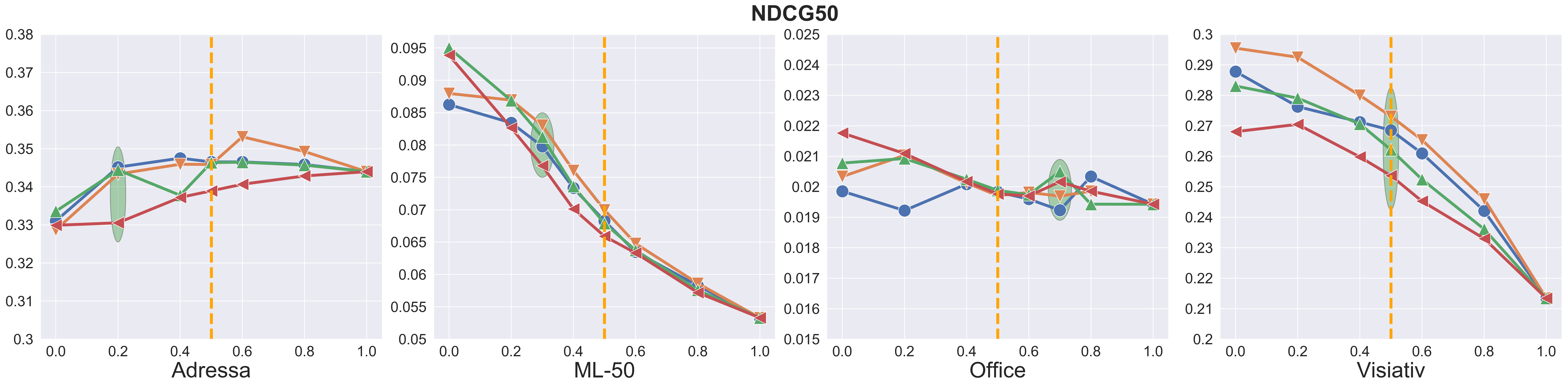}
	\caption{\modif{Influence on AUC, HIT\_50, NDCG\_50 (in ordinate) of $\gamma$ (in abscissa) for \algo{} \algo$_{1MC}$, \algo$_{2MC}$ and \algo$_{3MC}$.The green ellipse is the best $\gamma$ identified in \ref{sec:perf_study}.}}
	\label{fig:xp_gamma}
\end{figure}

The user preferences modelling part is very important to provide accurate recommendations. Models that only focus on user preferences (e.g., BPRMF, \algoLT{}) outperform those that only consider the sequential dynamics (e.g., FMC, \algoST{}, \algoST$_{1MC}$) on the AUC metric. However, there are also models that only focus on sequential dynamics and outperform those that only consider the user preferences on HIT\_X or NDGC\_X. The trade off between user preferences and sequential dynamics is very important to make \algo{} the most robust and accurate as possible for all kind of datasets and metrics. The $\gamma$ hyperparameter makes it possible to weight the importance given to each part of the model: \modif{$\gamma=0.5$} means that the two parts have the same importance, whereas the greater the $\gamma$, the greater the importance of the user preferences in the recommendations. \modif{The best values of $\gamma$ determined by grid search ($\gamma\in\lbrace 0.0,0.1,\dots,1.0\rbrace$) are reported in supplementary material \cite{REBUS_supplementary}. For most of datasets, $\gamma$ is between $0.3$ and $0.7$. This means that both user preferences and sequential dynamics are important, but one may dominate the other to get accurate recommendations. In Figure~\ref{fig:xp_gamma}, we analyze the impact of $\gamma$ on AUC, HIT\_50, NDCG\_50. We can observe different cases: (1)  $\gamma$ does not really have an impact on performance of \algo{} (e.g., Adressa dataset); (2) the best performance on AUC appears when \algo{} focused on the sequential dynamics (e.g., ML-50 dataset, $\gamma = 0.3$); (3) best results are obtained when \algo{} is focus on the user preferences (e.g., Office dataset, $\gamma = 0.7$); (4) the best performance on AUC is reached by \algo{} when the user preferences and the sequential dynamics have  the same importance (e.g., Visiativ). 
Notice that for Office dataset,   HIT\_50 and NDCG\_50 do not follow the same behavior as AUC: a greater $\gamma$ increases the AUC performance but decreases the HIT\_50 and NDCG\_50 performances. 
Eventually,  the green ellipse in Figure~\ref{fig:xp_gamma} is the best $\gamma$ identified for \algo{} in Section \ref{sec:perf_study} (we have selected the  hyperparameters that maximize the AUC performance) but we can see that if we change $\gamma$ for \algo$_{XMC}$ we can sometimes obtain better performance on AUC, HIT\_50, NDCG\_50 than those  reported on Tables \ref{tab:results} and \ref{tab:resultsbis}.}

\subsection{Examples of recommendations}
\begin{figure}[!htb]
	\includegraphics[width=\columnwidth]{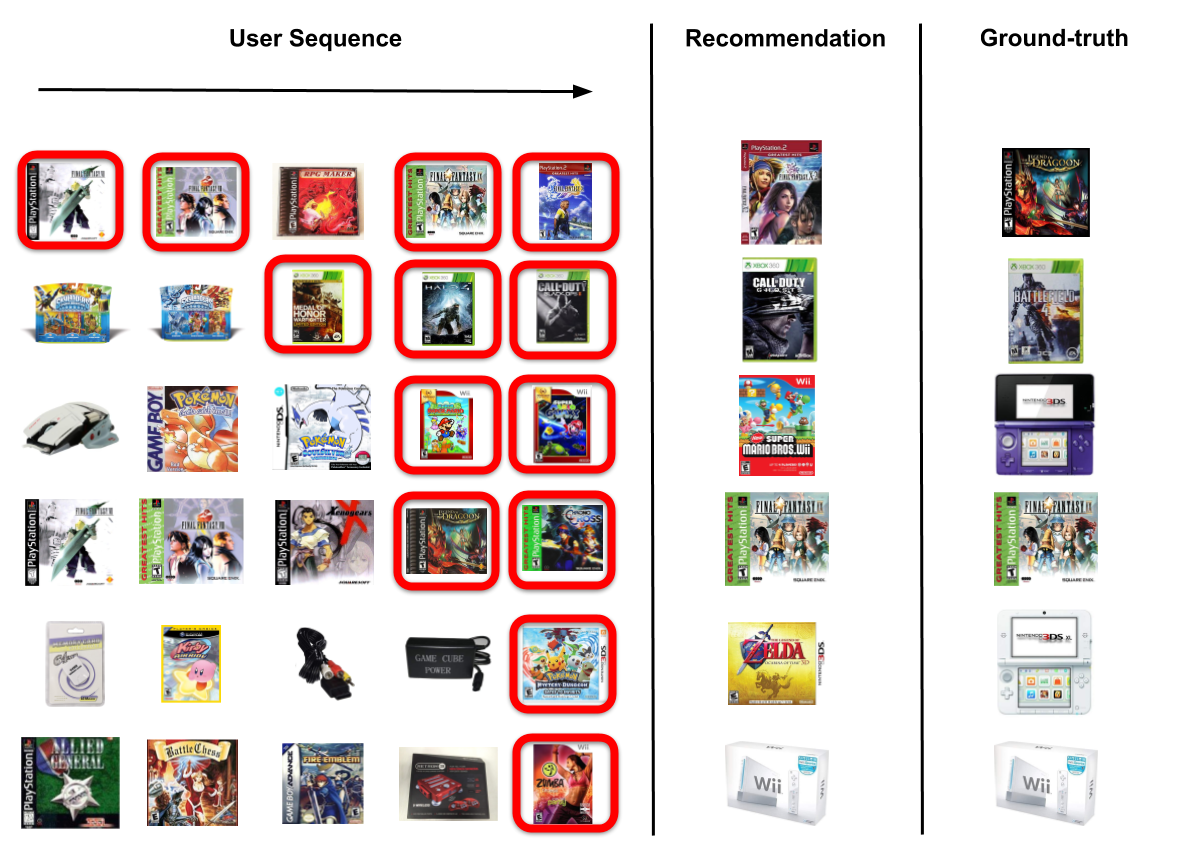}
	\caption{Recommendations  for a random sample of  users by \algo{} on Amazon Video Games. The red squares are items in $\frequ$. }
	\label{fig:size_histo_VS_size_fseq_histo_VS_nb_item_before}
\end{figure}

Figure~\ref{fig:size_histo_VS_size_fseq_histo_VS_nb_item_before} shows some recommendations made by \algo{} on Amazon-Video Games. As can be seen on these examples, \algo{} captures the sequential dynamics and recommends items which are similar to the ground truth. For instance, for the first user, \algo{} recommends \textit{Final Fantasy X-2} because the user sequentially bought the previous editions of the \textit{Final Fantasy} games (the red squared items). The ground truth is a similar game: \textit{The Legend of Dragoon}. On this example, the sequential dynamics is captured with a 4-item substring mapped with a discontinuity (a hole) on the user sequence. 
Some recommendations are based on smaller and more compact sequences, e.g., the two last recommendations use only the most recent item to capture the sequential dynamics. 
Based on the last user action --\textit{ Pokemon Mystery Dungeon on Nintendo 3DS} and \textit{Zumba Fitness on Nintendo Wii} -- \algo{} respectively recommends \textit{The Legend of Zelda: Ocarina of Time on Nintendo 3DS} (instead of \textit{Nintendo 3DS XL}) and a \textit{Nintendo Wii}.

\subsection{Summary and discussion}
This experimental study demonstrates that our sequential recommendation model \algo{} provides the most accurate recommendations for most of the datasets. Indeed \algo{} outperforms other methods in sparse data. When considering dense data, our model remains competitive and we have seen that the performances of \algo{} can even be increased with the use of shorter users' histories to model the sequential dynamics. For instance, \algoST{} -- that uses Markov chains of length 3 and the damping factor to model the sequential dynamics -- outperforms the state-of-the-art models on dense data (e.g., ML-50 or ML-30). In addition, we can use the substrings that support the recommendations to understand the mechanisms at work in this process. \modif{This empirical study also gives evidence that \algo{}  shows good performance for the problem of cold-start users outperforming most of state-of-the-art models.} Actually, the item embedding is powerful even if the history of the user is very short (e.g., 1 or 2 items). 
\modif{In  this work, $\gamma$ is a hyperparameter. An improvement could be to learned $\gamma$ during the learning phase.  We can also learn a new vector  $\Upsilon_u$ associated to each user to have a more personalized trade off between user preferences and sequential dynamics. This can be done, for instance,  by self attention model architecture \cite{SelfAttention}.
}

\section{Conclusion}\label{sec:con}
\algo{}, a new metric embedding model where items are only projected once, addresses the problem of sequential recommendation. It uses frequent substrings as a proxy to learn conditional distributions of item appearance in the sequential history of the user in order to provide personalized order Markov chains and thus capture sequential dynamics. Our model learns a representation of the user preferences and the sequential dynamics in the same Euclidean space. We have carried out thorough experiments on multiple real-world datasets. These experiments demonstrate that \algo{} outperforms all the state-of-the-art models on sparse datasets -- which is the most standard configuration for recommender systems -- and has comparable performances to those of state-of-the-art models for dense datasets.
The sequences used to model the sequential dynamics also provide additional insights with the recommendations. This work opens up several avenues for further research such as the investigation of more sophisticated pattern mining techniques to both better model the user preferences and the sequential dynamics, and to provide explanations on the recommendations.

\subsection*{\textbf{Acknowledgements.}}
This work was supported by
the ACADEMICS grant of the IDEXLYON, project of the University of Lyon, PIA
operated by ANR-16-IDEX-0005.




\bibliographystyle{unsrt}
\bibliography{biblio}
	
\end{document}